\documentclass[10pt,twocolumn,letterpaper]{article}

\usepackage{iccv}
\usepackage{times}
\usepackage{epsfig}
\usepackage{graphicx}
\usepackage{amsmath}
\usepackage{amssymb}
\usepackage{booktabs}
\usepackage{multirow}
\usepackage{color}
\usepackage{colortbl}  %彩色表格需要加载的宏包
\usepackage{xcolor}
\usepackage{array}  %对表列和表格线的设置需要用到array宏包
\usepackage{algorithm}
\usepackage{algorithmicx}
\usepackage{algpseudocode}
\usepackage{makecell}
\definecolor{tableColor}{HTML}{e9f1f6} 

\usepackage[pagebackref=true,breaklinks=true,letterpaper=true,colorlinks,bookmarks=false]{hyperref}

\iccvfinalcopy % *** Uncomment this line for the final submission

\def\iccvPaperID{4815} % *** Enter the ICCV Paper ID here

\ificcvfinal\fi

\begin{document}

%%%%%%%%% TITLE
\title{Focus the Discrepancy: Intra- and Inter-Correlation Learning for Image Anomaly Detection}

% \and
\author{
Xincheng Yao$^{{\rm 1}}$, Ruoqi Li$^{{\rm 1}}$, Zefeng Qian$^{{\rm 1}}$, Yan Luo$^{{\rm 1}}$, Chongyang Zhang$^{{\rm 1},{\rm 2}}\thanks{Corresponding Author.}$\\
\textsuperscript{\rm 1}School of Electronic Information and Electrical Engineering, Shanghai Jiao Tong University\\
\textsuperscript{\rm 2}MoE Key Lab of Artificial Intelligence, AI Institute, Shanghai Jiao Tong University\\
{\tt\small \{i-Dover, nilponi, zefeng\_qian, luoyan\_bb, sunny\_zhang\}@sjtu.edu.cn$^{{\rm 1}}$}
}

\maketitle
\ificcvfinal\thispagestyle{empty}\fi

%%%%%%%%% ABSTRACT
\begin{abstract}
    Humans recognize anomalies through two aspects: larger patch-wise representation discrepancies and weaker patch-to-normal-patch correlations. However, the previous AD methods didn't sufficiently combine the two complementary aspects to design AD models. To this end, we find that Transformer can ideally satisfy the two aspects as its great power in the unified modeling of patch-wise representations and patch-to-patch correlations. In this paper, we propose a novel AD framework: FOcus-the-Discrepancy (FOD), which can simultaneously spot the patch-wise, intra- and inter-discrepancies of anomalies. The major characteristic of our method is that we renovate the self-attention maps in transformers to Intra-Inter-Correlation (I2Correlation). The I2Correlation contains a two-branch structure to first explicitly establish intra- and inter-image correlations, and then fuses the features of two-branch to spotlight the abnormal patterns. To learn the intra- and inter-correlations adaptively, we propose the RBF-kernel-based target-correlations as learning targets for self-supervised learning. Besides, we introduce an entropy constraint strategy to solve the mode collapse issue in optimization and further amplify the normal-abnormal distinguishability. Extensive experiments on three unsupervised real-world AD benchmarks show the superior performance of our approach. Code will be available at \url{https://github.com/xcyao00/FOD}.
\end{abstract}

%%%%%%%%% BODY TEXT
\section{Introduction}
\label{sec:intro}

\begin{figure}
    \centering
    \includegraphics[width=1.0\linewidth]{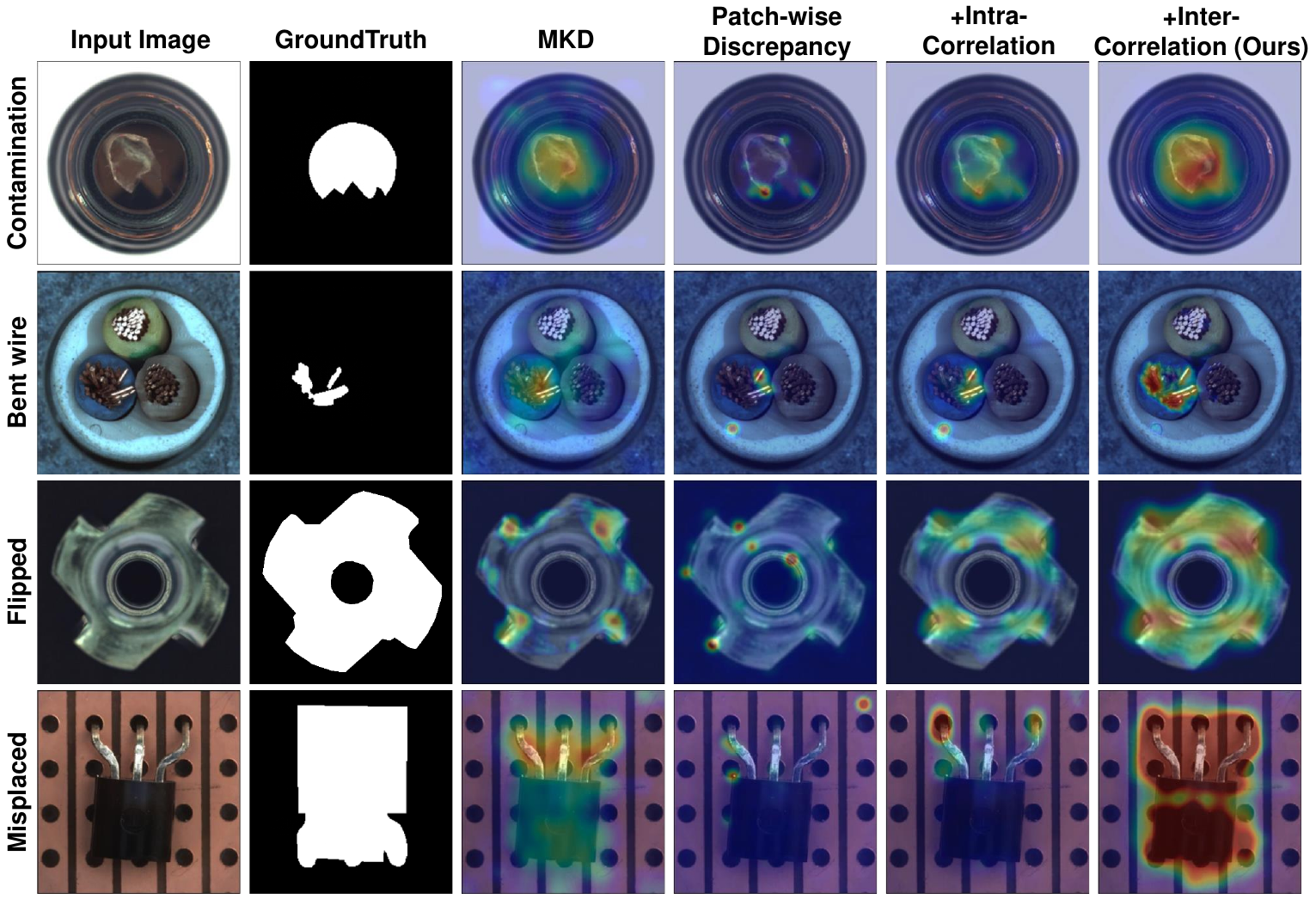}
    \caption{Anomaly detection examples on MVTecAD \cite{MVTecAD}. Multiresolution Knowledge Distillation (MKD) \cite{MKD} adopts the conventional patch-wise representation discrepancies. Row 1 shows the hard global anomalies (\emph{i.e}, they are not significantly different from normal visuals). Rows 3 and 4 show the logical anomalies (\emph{i.e.} they may be easily recognized as normal if only from the patch-wise discrepancy).}
    \label{fig:motivation}
\end{figure}

The goal of anomaly detection (AD) is to distinguish an instance containing anomalous patterns from those normal samples and further localize those anomalous regions. Anomalies are defined as opposite to normal samples and are usually rare, which means that we need to tackle AD tasks under the unsupervised setting with only normal samples accessible. The core idea of most unsupervised AD methods is to compare with normal samples to distinguish anomalies \cite{Review1, PatchCore, PatchSVDD, PaDiM, MemoryAE, BGAD}. Even for humans, we also recognize anomalies in this way, specifically through three discrepancies, \emph{i.e.}, 1. patch patterns that differentiate from the normal visuals; 2. image regions that destroy textures or structures; 3. novel appearances that deviate from our accumulated knowledge of normality. Namely, anomalous patches usually have three characteristics: their patch-wise representations are different from the normal visuals; they are different from most patches within one image; they deviate from our accumulated knowledge of normality. These views intrinsically reveal that humans' recognition of anomalies depends on two aspects: patch-wise representations (1) and intra- and inter-image correlations (2, 3). 

% \emph{i.e.}, anomalies are different from normal visual from the local view, and their correlations with most patches are weak
 
Previous methods mainly follow the former aspect to learn distinguishable representations or reconstructions, such as reconstruct-based methods \cite{SSIM, DFR, GANomaly} and knowledge distillation AD models \cite{STAD, MKD}. The goal of these methods is to generate reconstructed samples or feature representations, and larger patch-wise representation discrepancies can appear in the abnormal patches. However, only the patch-wise representation discrepancies are insufficient for detecting more complex anomalies (\emph{e.g.}, rows 3 and 4 in Figure \ref{fig:motivation}), since the patch-wise errors can't provide comprehensive descriptions of the spatial context. Other mainstream AD methods, such as embedding-based \cite{PaDiM, PatchCore} and one-class-classification-based (OCC) \cite{deepSVDD, PatchSVDD} methods, are much similar to the latter aspect. These methods achieve anomaly detection by measuring the distances between the features of test samples and normal features. Compared with the non-learnable feature distances, the explicit intra- and inter-image correlations in our method are more effective to detect diverse anomalies (see Table \ref{tab:MVTecAD}, \ref{tab:detailed_MVTecAD}, \ref{tab:BTAD_MVTec3D}). Moreover, patch-wise representation discrepancies and intra- and inter-correlation discrepancies are complementary, and can be combined to develop more powerful AD models.

Recently, with the self-attention mechanism and long-range modeling ability, transformers \cite{Transformer} have significantly renovated many computer vision tasks \cite{ViT, SwinTransformer, DETR, SETR, SegFormer} and recently popular language-vision multimodal tasks \cite{CLIP, GLIP}. Transformers have shown great power in the unified modeling of patch-wise representations and patch-to-patch correlations. Transformers are quite suitable for AD tasks as their modeling ability can satisfy the two aspects of anomaly recognition quite well. Some works \cite{BTAD, Intra, UTRAD, UniAD} also attempt to employ transformers to construct AD models. However, these methods only use transformers to extract vision features, which didn't sufficiently adapt transformers' long-range correlation modeling capability to AD tasks. Different from these works, we explicitly exploit transformers' self-attention maps to establish the intra- and inter-image correlations. The correlation distribution of each patch can provide more informative descriptions of the spatial context, which can reveal more intricate and semantic anomaly patterns.

In this paper, motivated by humans' anomaly recognition process, we propose a novel
AD framework: FOcus-the-Discrepancy (FOD), which can
exploit transformers' unified modeling ability to simultaneously spot the patch-wise, intra- and inter-discrepancies. Our key designs are composed of three recognition branches: the patch-wise discrepancy branch is to reconstruct the input patch features for distinguishing simple anomalies; the intra-correlation branch is to explicitly model patch-to-patch correlations in one image for distinguishing hard global anomalies (\emph{e.g.}, row 1 in Figure \ref{fig:motivation}); the inter-correlation branch is to explicitly learn inter-image correlations with known normal patterns from the whole normal training set. To implement the intra- and inter-correlation branches, we adapt Transformer and renovate the self-attention mechanism to the I2Correlation, which contains a two-branch structure to first separately model the intra- and inter-correlation distribution of each image patch, and then fuse the features of two-branch to spotlight the abnormal patterns. To learn the intra- and inter-correlations adaptively, we propose the RBF-kernel-based target-correlations as learning targets for self-supervised learning, the RBF kernel is used to present the neighborhood continuity of each image patch. Besides, an entropy constraint strategy is applied in the two branches, which can solve the mode collapse issue in optimization and further amplify the normal-abnormal distinguishability.

In summary, we make the following main contributions:

1. We propose a novel AD framework: FOD, which can effectively detect anomalies by simultaneously spotting the patch-wise, intra- and inter-discrepancies.

2. We renovate the self-attention mechanism to the I2Correlation, which can explicitly establish intra- and inter-correlations in a self-supervised way with the target-correlations. An entropy constraint strategy is proposed to further amplify the normal-abnormal distinguishability.

3. Our method can achieve SOTA results on three real-world AD datasets, this shows our method can more effectively determine anomalies from complementary views.

%-------------------------------------------------------------------------
\section{Related Work}

\textbf{Anomaly Detection.} In this paper, we divide the mainstream AD methods into five categories: reconstruction-based, embedding-based, OCC-based methods, knowledge distillation and normalizing flow AD models. The reconstruction-based methods are the most popular AD methods and also widely studied, where the assumption is that models trained by normal samples only can reconstruct normal regions but fail in abnormal regions. Many previous works attempt to train AutoEncoders \cite{SSIM, MemoryAE, DFR, UTRAD, DivideAssemble, DRAEM}, Variational AutoEncoders \cite{VAE1} and GANs \cite{AnoGAN, GANomaly, GAN1, GAN2} to reconstruct the input images. Overfitting to the input images is the most serious issue of these methods, which means that the anomalies are also well reconstructed \cite{UniAD}. 

 Recently, some embedding-based methods \cite{DeepKNN, GaussianAD, SPADE, PaDiM, PatchCore} show better AD performance by using ImageNet pre-trained networks as feature extractors. In \cite{GaussianAD}, the authors fit a multivariate Gaussian to model the image-level features for further Mahalanobis distance measurement. PaDiM \cite{PaDiM} extends the above method to localize pixel-level anomalies. PatchCore \cite{PatchCore} extends on this line by utilizing locally aggregated, mid-level features and introducing greedy coreset subsampling to form nominal feature banks. However, these methods directly utilize pre-trained networks without any adaptation to the target dataset. Some works \cite{PANDA, TransferAD, MeanShift} attempt to adapt pre-trained features to the target data distribution. There are also some other methods for using pre-trained networks by freezing them and only optimizing a sub-network, \emph{e.g.}, via knowledge distillation \cite{STAD, MKD, RDAD, AST}, or normalizing flows \cite{DifferNet, CFLOW, CS-FLOW, FastFlow}. 

 OCC is another classical AD paradigm, the earliest works are mainly to extend the OCC models such as OC-SVM \cite{OneclassSVM} or SVDD \cite{SVDD, deepSVDD} for anomaly detection. Recently, in \cite{PatchSVDD}, a patch-based SVDD that contains multiple cores rather than a single core in \cite{deepSVDD} is proposed to enable anomaly localization. In \cite{FCDD}, a Fully Convolutional Data Description combined with receptive field upsampling is proposed to generate anomaly maps. In \cite{MS-PatchSVDD}, the authors further extend the PatchSVDD \cite{PatchSVDD} model by the proposed multi-scale patch-based representation learning method.

\textbf{Transformer-based Anomaly Detection Methods.} Recently, transformers \cite{Transformer} have shown great power in modeling long-range dependencies. For image anomaly detection, some works \cite{BTAD, Intra, UniAD, UTRAD, PMAD} also attempt to exploit transformers to design AD models. However, most of these methods only use transformers as backbones to extract vision features, and don't effectively adapt the long-range modeling capacity of transformers to AD tasks. Unlike the previous usage of transformers, we explicitly exploit the self-attention maps of transformers to establish intra- and inter-correlations. Our work shares some similarities with a recent work \cite{Anomaly-Transformer}. But we point out that our work has some significant differences from \cite{Anomaly-Transformer}: 1) Different Insight: \cite{Anomaly-Transformer} aims to unify the pointwise representation and pairwise association for time series AD, whereas our work aims to more sufficiently detect image anomalies through three complementary recognition views. 2) Novel Method: we propose entropy constraint, inter-correlation branch, and I2Correlation by effectively combining intra- and inter-correlation branches. 3) Different Task: \cite{Anomaly-Transformer} focuses on time-series AD, whereas we focus on image AD.

%------------------------------------------------------------------------

\section{Approach}
\label{sec:approach}

\begin{figure*}[ht]
    \centering
    \includegraphics[width=1.0\linewidth]{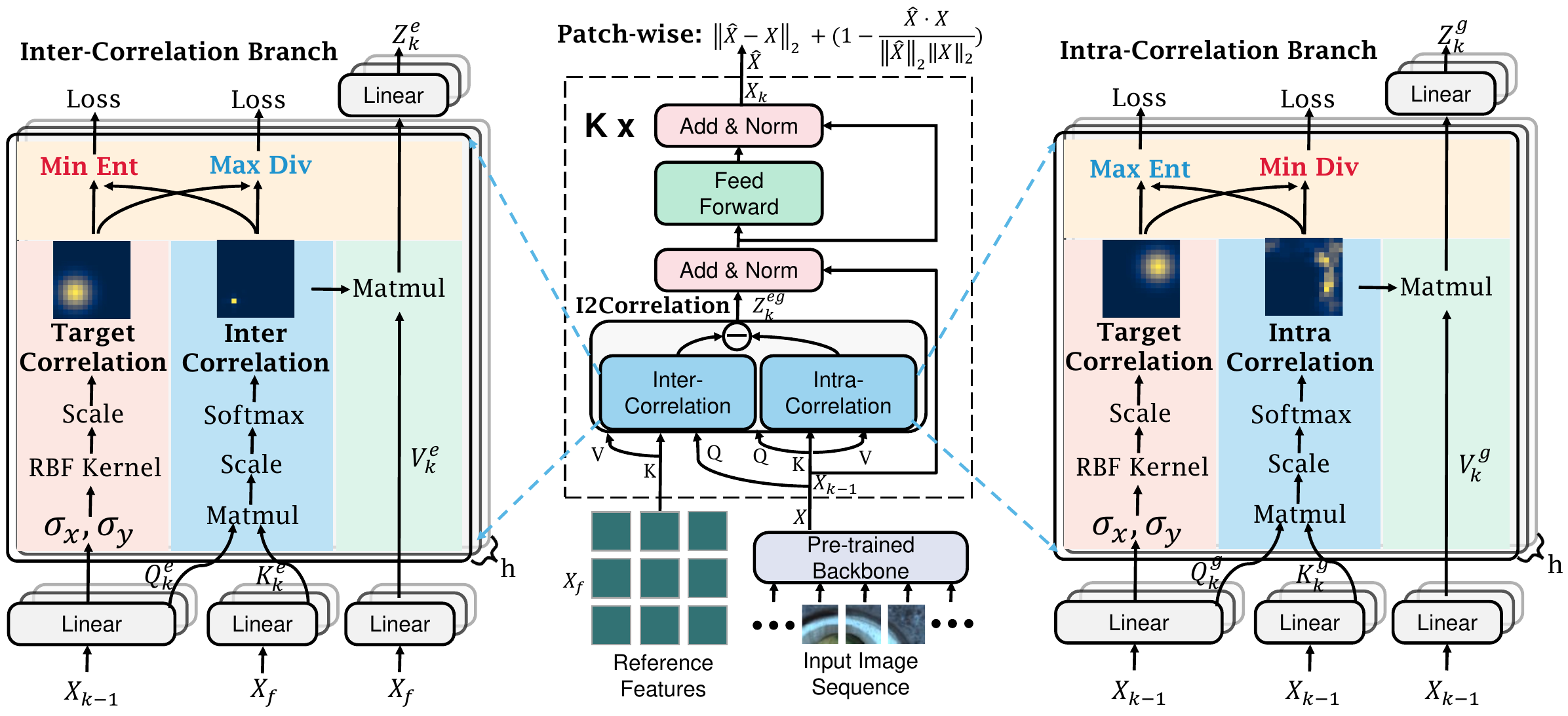}
    \caption{Model overview. The model is composed of three parts: patch-wise discrepancy branch, intra-correlation branch, and inter-correlation branch. The input image patch sequence will be transformed into the hidden features by a pre-trained backbone, and then sent into the intra-correlation branch for learning intra-image patch-to-patch correlations. The input of the inter-correlation branch is composed of hidden features (as Query) and reference features (as Key and Value). The final output features are used to reconstruct the input features.}
    \label{fig:framework}
\end{figure*}

\subsection{Model Overview}
\label{sec:network_architecture}
 Figure \ref{fig:framework} overviews our proposed approach. The model consists of three branches: patch-wise discrepancy branch, intra-correlation branch, and inter-correlation branch. The input 2D image is first sent into a pre-trained backbone to extract multi-scale feature maps. At each feature level, we construct a subsequent transformer network to explicitly model the intra- and inter-image correlations for spotting the intra- and inter-discrepancies. Each network is implemented by stacking the I2Correlation blocks and feed-forward layers alternately. Suppose each network contains $K$ layers with length-$N$ input features $X \in \mathbb{R}^{N \times d}$, the output of $k$th layer $X_k \in \mathbb{R}^{N \times d_{m}}$ is calculated as follows:
 \begin{align}
     &Z_k = {\rm LN}({\rm I2Correlation}(X_{k-1}, X_f) + X_{k-1}) \\
     &X_k = {\rm LN}({\rm FeedForward}(Z_k)+Z_k)
 \end{align}
 where $X_f$ is the reference features used by the inter-correlation branch, $LN$ means LayerNorm, and $Z_k \in \mathbb{R}^{N \times d_{m}}$ is the $k$th layer's hidden features. The final output features are calculated by linear projection: $\hat{X} = Z_KW_o$, where $W_o \in \mathbb{R}^{d_{m} \times d}$ is the output projection matrix. 

 % flattened into 1D patch sequence $I_p \in \mathbb{R}^{N \times (P^2\times C)}$, where $(P, P)$ is the resolution of each image patch and $N = HW / P^2$ is the number of patches, and then
 
 \subsection{Patch-Wise Reconstruction Discrepancy}
 Adopted from previous reconstruction-based AD methods, we employ feature reconstruction as our patch-wise recognition view for simplicity. In our approach, we construct a transformer network to reconstruct the input features. With the long-range dependency modeling ability, the features reconstructed by Transformer can have larger effective receptive fields \cite{RepLKNet}, which are more conducive to detect hard global anomalies and logical anomalies. Moreover, with the further introduced inter-correlation branch (sec.\ref{sec:external_view}), the features can even perceive normal regions of images from the whole normal training set. Therefore, the reconstructed features generated by our model have better global perception and more discriminative semantic representation capability, which are more suitable for anomaly detection by patch-wise representation discrepancies.

 \textbf{Learning Objective.} We can utilize classical reconstruction losses as the learning objective. We combine the $\ell2$ distance and cosine distance to measure the feature distances between the reconstructed features $\hat{X}$ and the input features $X$. The loss function is defined as follows:
 \begin{equation}
 \label{eq:local_loss}
     \mathcal{L}_{l} = ||\hat{X} - X||_2 + \Bigg(1 - \frac{\hat{X}\cdot X}{||\hat{X}||_2||X||_2}\Bigg)
 \end{equation}
 
 \subsection{Intra-Correlation Learning}
   The intra-correlation learning branch aims to learn informative patch-to-patch correlations from the input patch sequence adaptively. As shown in the right part of Figure \ref{fig:framework}, we explicitly take advantage of the self-attention maps of transformers as intra-correlation matrices. Formally, the intra-correlation matrix of the $k$th layer is calculated by:
   \begin{align}
       &{\rm Intra \; Correlation}: S^g_k = {\rm Softmax}\Big(Q^g_k(K^g_k)^T/\sqrt{d_{m}}\Big) \nonumber \\
       & [Q^g_k, \; K^g_k, \; V^g_k] = X_{k-1}[W_k^{Q_g}, \; W_k^{K_g}, \; W_k^{V_g}]
   \end{align}
   where $Q^g_k, K^g_k, V^g_k \in \mathbb{R}^{N \times d_{m}}$ represent the query, key, and value of the $k$th layer. $W_k^{Q_g}, W_k^{K_g}, W_k^{V_g} \in \mathbb{R}^{d_{m} \times d_{m}}$ represent the learnable projection matrices for $Q^g_k, K^g_k, V^g_k$, respectively. $S^g_k \in \mathbb{R}^{N \times N}$ denotes the learned intra-image patch-to-patch correlations. Since ${\rm Softmax}(\cdot)$ can convert the values in the similarity map into range $[0,1]$ along the horizontal axis, each row of $S^g_k$ can represent a discrete correlation distribution for each corresponding image patch. But different from the vanilla transformers, we further introduce target correlation matrices as the learning objective to explicitly optimize the intra-image correlations.
   
   \textbf{Target Correlation.} Inspired by the contrastive learning method BarlowTwins \cite{BarlowTwins}, we can construct a target correlation matrix $T \in \mathbb{R}^{N \times N}$ as the learning target. The objective function of BarlowTwins measures the correlation matrix between the embeddings of two identical networks fed with distorted versions of a batch of samples, and tries to make this matrix close to the identity. In our work, the role of the target correlation is to introduce a prior correlation of patches as a pretext learning target, where each patch can be highly correlated to itself and also correlated to its neighborhood patches and the correlation decreases with the increase of distance. This allows us to optimize the intra- and inter-correlations in a self-supervised way. To this end, we use the radial basis function (RBF) to construct the target correlation matrix. We further adopt two learnable kernel variances $\sigma_x$ and $\sigma_y$ for horizontal and vertical axes to make the target correlation of each patch can adapt to the specific pattern of itself. The target correlation matrix of the $k$th layer is defined as:
 \begin{align}
      & T^g_k = \frac{1}{2\pi\sigma_x\sigma_y}{\rm exp}\Big(-\frac{||x_{ij}-x_{i^\prime j^\prime}||_2^2}{2(\sigma_x^2+\sigma_y^2)}\Big) \nonumber \\
      & i, i^\prime \in \{1,\dots,H\}; j, j^\prime \in \{1,\dots,W\}
 \end{align}
 where $||x_{ij}-x_{i^\prime j^\prime}||_2^2$ means the Square Euclidean distance between point $x_{ij}$ and $x_{i^\prime j^\prime}$, $i, i^\prime$ and $j, j^\prime$ means vertical and horizontal coordinates, respectively. 
 
 Next, we need to measure the distance between target- and intra-correlation distributions. This can usually be achieved by calculating the KL divergence. We can obtain a KL divergence value from each level of the network. Thus, we average all KL divergence values to combine the patch-to-patch correlations from multi-layer features into a more informative measure as follows:
 \begin{equation}
 \label{eq:KL_Div}
     {\rm Div}(\mathcal{T}^g, \mathcal{S}^g) \!=\! \frac{1}{K}\sum_{k=1}^{K}\Big(KL(T_k^{g}||S_k^{g}) \!+\! KL(S_k^{g}||T_k^{g})\Big)
 \end{equation}
 where $KL(\cdot||\cdot) \in \mathbb{R}^N$ and its each element means the KL divergence between two discrete distributions corresponding to each row of $T^g_k$ and $S^g_k$. 
 
   Due to the rarity of anomalies and the dominance of normal patterns, the normal patches should build strong correlations with most patches in the whole image, while the weights of abnormal correlation distributions are harder to distribute to most patches and are more likely to concentrate on the adjacent image patches due to the neighborhood continuity. Since normal and abnormal patches have different correlation distributions, this is a distinguishable criterion for anomaly detection. Note that the intra-correlation branch explicitly exploits the spatial dependencies of each image patch, which are more informative than the patch-wise representations for anomaly detection. 

   \textbf{Entropy constraint.} Since the normal image patterns are usually diverse, the learned correlation distributions of the normal patches may also easily concentrate on the adjacent patches, which can cause the distinguishability between normal and abnormal to be downscaled. To address this, we further introduce an entropy constraint item for making normal patches establish strong associations with most normal patches in the whole image as much as possible. The entropy constraint item is defined as:
 \begin{equation}
     {\rm Ent}(\mathcal{S}^g) \!=\! \frac{1}{K}\sum_{k=1}^{K}\Big(\sum_{i=1}^{N}\sum_{j=1}^N(-S^g_{k(i,j)}{\rm log}(S^g_{k(i,j)}))\Big)
 \end{equation}
  
  We will maximize the entropy constraint item. The loss function for the intra-correlation branch is defined as:
 % \begin{equation}
 % \label{eq:loss_global}
 %     \mathcal{L}_{g} = \lambda_1{\rm Div}(\mathcal{T}_g, {\rm SG}[\mathcal{S}_g]) - \lambda_1{\rm Div}({\rm SG}[\mathcal{T}_g], \mathcal{S}_g) - \lambda_2{\rm Ent}(\mathcal{S}_g)
 % \end{equation}
 \begin{equation}
 \label{eq:loss_global}
     \mathcal{L}_{g} = \lambda_1{\rm Div}(\mathcal{T}^g, \mathcal{S}^g)  - \lambda_2{\rm Ent}(\mathcal{S}^g)
 \end{equation}
 where $\lambda_1$ and $\lambda_2$ are used to trade off the loss items. The optimization of intra-correlation $\mathcal{S}^g$ is actually an alternating process with the guidance of target correlation $\mathcal{T}^g$ (see App. \ref{sec:sup_opt}), ultimately resulting that each normal patch can establish strong correlations with most normal patches.

 \subsection{Inter-Correlation Learning}
 \label{sec:external_view}
 Through the intra-correlation branch, we can establish patch-to-patch correlations within a single image. However, an image usually doesn't contain all possible normal patterns, which may cause it difficult to distinguish some ambiguous abnormal patches (see rows 2, 3, 4 in Figure \ref{fig:motivation}) only through the intra-correlations. To address this, we should effectively take advantage of the known normal patterns from the normal training set, which are more likely to contain more informative normal patterns. Specifically, we further propose an inter-correlation learning branch to explicitly model pairwise correlations with the whole normal training set. In this branch, the features of each patch establish a discrete inter-correlation distribution with the reference features extracted from all normal samples (see Figure \ref{fig:framework}). The inter-correlation matrix of the $k$th layer is similar to the corresponding intra-correlation, and is defined as:
   \begin{align}
       & {\rm Inter \; Correlation}: S^e_k = {\rm Softmax}\Big(Q^e_k(K^e_k)^T/\sqrt{d_{m}}\Big)  \nonumber \\
       & Q^e_k, \;\; K^e_k, \;\; V^e_k = X_{k-1}W_k^{Q_e}, \;\; X_{f}W_k^{K_e}, \;\; X_{f}W_k^{V_e}
   \end{align}
where $X_f \in \mathbb{R}^{N_e \times d_e} $ represents the external reference features, $N_e$ is the length of the reference features and $d_e$ is the feature dimension. $W_k^{Q_e} \in \mathbb{R}^{d_{m} \times d_{m}}$ and  $W_k^{K_e}, W_k^{V_e} \in \mathbb{R}^{d_{e} \times d_{m}}$ are learnable matrices for $Q^e_k, K^e_k, V^e_k$. $S^e_k \in \mathbb{R}^{N\times N_e}$ denotes the learned inter-image correlations. 
%Each row of $S^e_k$ can represent a discrete external correlation distribution for each corresponding image patch.

\textbf{Loss.} The loss function for the inter-correlation branch has the opposite optimization direction to $\mathcal{L}_g$. Because the external reference features contain more comprehensive normal patterns, normal patches can establish stronger correlations with the closest reference normal patterns, instead of establishing strong correlations with most reference patterns. By contrast, it shall be harder for anomalous patches to establish strong correlations with any of the reference patterns. So the inter-correlation distributions of anomalies are more dispersed, while the normal inter-correction distributions are more likely to be concentrated. To this end, we maximize the KL divergence and minimize the entropy item in the training process. The practical optimization strategy is also opposite to the intra-correlation branch (see App. \ref{sec:sup_opt} for details). The loss function is defined as:
 % \begin{equation}
 %  \label{eq:loss_external}
 %     \mathcal{L}_{e} = -\lambda_1{\rm Div}(\mathcal{T}_e, {\rm SG}[\mathcal{S}_e]) + \lambda_1{\rm Div}({\rm SG}[\mathcal{T}_e], \mathcal{S}_e) + \lambda_2{\rm Ent}(\mathcal{S}_e)
 % \end{equation}
 \begin{equation}
  \label{eq:loss_external}
     \mathcal{L}_{e} = -\lambda_1{\rm Div}(\mathcal{T}^e, \mathcal{S}^e) + \lambda_2{\rm Ent}(\mathcal{S}^e)
 \end{equation}
 
\textbf{External Reference Features.} External reference features are used for providing accumulated knowledge of normality for the inter-correlation learning branch. Thus, these features should represent all possible normal patterns of all normal samples from the whole normal training set. To this end, we can employ many methods to generate the reference features, such as sampling key features by coreset subsampling algorithm \cite{PatchCore}, generating prototype features by memory module \cite{MemoryAE}, or learning codebook features through vector quantization \cite{DSR} or sparse coding techniques \cite{Sparse}. However, because the RBF-kernel in $\mathcal{T}^e$ is position-sensitive, our reference features are better to preserve the positional information. The ablation results in App.\ref{sec:sup_ablation} show that the methods that can't preserve the position information perform worse. From comprehensive ablation studies, we find that using patch-wise averaged features as the external reference features is a simple but effective way. We think that features extracted by deep neural networks are highly redundant \cite{GhostNet, TBC}, and different normal patterns generally correspond to larger activation values at different channels in the feature vector \cite{NeuralVis}. So feature averaging will not eliminate some rare normal patterns \cite{PaDiM}, these patterns may be preserved at specific channels. And averaging can greatly reduce the feature redundancy, making the obtained reference features more representative. Formally, for position $(i, j)$, we first extract the set of patch features at $(i, j)$, $X_{ij} = \{x_{ij}^{k}\}, k \in [1, N]$ from the $N$ normal training images. Then, the reference features at position $(i, j)$ is computed as $x_{ij}^f = \frac{1}{N}\sum_{k=1}^{N}x_{ij}^k$. The final external reference features are composed of averaged features at all locations and then flattened into 1D: $X_f = {\rm Flatten}(\{x_{ij}^{f}\})$.

 \subsection{I2Correlation}
 \label{sec:GECorrelation}

  We further combine the intra-correlation and inter-correlation branches to form the I2Correlation block. The output features of the intra- and inter-correlation branches are defined as: $Z^g_k = S^g_kV^g_k$ and $Z^e_k = S^e_kV^e_k$, respectively. Then, we use the residual feature $Z^{ge}_k = Z^g_k - Z^e_k$ as the output of the I2Correlation block. The feature $Z^e_k$ is generated from the external reference features, which can contain rich normal patterns. Thus, by subtracting $Z^e_k$ from the feature $Z^g_k$, it is conducive to spotlight the abnormal patterns in the $Z^g_k$. This is beneficial for anomaly detection.
 
 The total loss function consists of the there branch loss functions, and is combined as follows:
 \begin{equation}
     \mathcal{L}_{total} = \mathcal{L}_{l} + \mathcal{L}_{g} + \mathcal{L}_{e}
 \end{equation}
 
 \subsection{Anomaly Scoring}
 We utilize reconstruction errors as the baseline anomaly criterion and incorporate the normalized correlation distribution distances into the reconstruction criterion. The final anomaly score of the $i$th patch in the input patch sequence is shown as follows:
 % \begin{align}
 % \label{eq:combined}
 %      & S(I_p) = \Big[||X^0_{i,:} - \hat{X}_{i,:}||_2 + \big(1 - \frac{X^0_{i,:} \cdot \hat{X}_{i,:}}{||X^0_{i,:}||_2||\hat{X}_{i,:}||_2}\big)\Big]\odot \\ 
 %     &{\rm Softmax}\big(-{\rm Div}(\mathcal{T}_g, \mathcal{S}_g)\big)\odot{\rm Softmax}\big(-{\rm Div}(\mathcal{T}_e, \mathcal{S}_e)\big) \nonumber
 % \end{align}
  \begin{align}
 \label{eq:combined}
      & s_i = \Big[||X_{i,:} - \hat{X}_{i,:}||_2 + \big(1 - \frac{X_{i,:} \cdot \hat{X}_{i,:}}{||X_{i,:}||_2||\hat{X}_{i,:}||_2}\big)\Big]\odot \\ 
     &\quad\quad\quad\quad\quad\quad \Big(1 - {\rm Softmax}\big(-{\rm Div}(\mathcal{T}^e_{i,:}, \mathcal{S}^e_{i,:})\big)\Big) \nonumber
 \end{align}
 where $\odot$ is the element-wise multiplication.

%------------------------------------------------------------------------
\section{Experiments}
\label{sec:experiments}

\subsection{Experimental Setup}
\label{sec:exp_dataset}
\textbf{Datasets.} We extensively evaluate our approach on two widely used industrial AD datasets: the MVTecAD \cite{MVTecAD} and BTAD \cite{BTAD}, and one recent challenging dataset: the MVTecAD-3D \cite{MVTecAD3D}. MVTecAD is established as a standard benchmark for evaluating unsupervised image anomaly detection methods. This dataset contains 5354 high-resolution images from 15 real-world categories. 5 classes consist of textures and the other 10 classes contain objects. A total of 73 different anomaly types are presented. BTAD is another popular benchmark for unsupervised image anomaly detection, which contains 2540 RGB images of three industrial products. All classes in this dataset belong to textures. MVTecAD-3D is recently proposed, which contains 4147 2D RGB images paired with high-resolution 3D point cloud scans from 10 real-world categories. Even though this dataset is mainly designed for 3D anomaly detection, most anomalies can also be detected only through RGB images without 3D point clouds. Since we focus on image anomaly detection, we only use RGB images of the MVTecAD-3D dataset. We refer to this subset as MVTec3D-RGB. 

\textbf{Evaluation Metrics.} The standard metric in anomaly detection, AUROC, is used to evaluate the performance of AD methods. Image-level AUROC is used for anomaly detection and a pixel-based AUROC for evaluating anomaly localization \cite{MVTecAD, STAD, PaDiM, DRAEM}.

\textbf{Implementation Details.} We use EfficientNet-b6 \cite{Efficientnet} to extract two levels of feature maps with $\{8\times, 16\times\}$ downsampling ratios, the pre-trained networks are from the timm library \cite{timm}. Then, we construct a subsequent transformer network (see Figure \ref{fig:framework}) at each feature level to reconstruct patch features and learn patch-to-patch correlations. The parameters of the feature extractor are frozen in the training process, only the parameters of the subsequent transformer networks are learnable. All the subsequent transformer networks in our model contain 3 layers. We set the hidden dimension $d_{m}$ as $\{256, 512\}$ and the number of heads as $8$. The hyperparameters $\lambda_1$ and $\lambda_2$ are set as $0.5$ and $0.5$ to trade off two parts of the $\mathcal{L}_g$ and $\mathcal{L}_e$ loss functions (see App. \ref{sec:sup_ablation} for hyperparameter sensitivity analysis). We use the Adam \cite{Adam} optimizer with an initial learning rate of $1{\rm e}^{-4}$. The total training epochs are set as 100 and the batch size is 1 by default. All the training and test images are resized and cropped to $256\times256$ resolution from the original resolution. 
%All the experiments are implemented in Pytorch \cite{Pytorch} with a single NVIDIA RTX 3090 GPU.

\begin{table}
\caption{Comparison of our method with the SOTA methods on the MVTecAD dataset. \textcolor{red}{Red} and \textcolor{blue}{blue} indicate the first and the second best result, respectively. According to the anomaly recognition process, we divide these methods into the patch-wise representation discrepancy, patch-to-patch feature distance, and others.}
\label{tab:MVTecAD}
\resizebox{1.0\linewidth}{!}{
\begin{tabular}{c||cccc}
\hline
\makecell[c]{Discrepancy \\ Type} & Method & Venue & \makecell[c]{Image-level \\ AUROC} & \makecell[c]{Pixel-level \\ AUROC}\\
\hline
\hline
 \multirow{13}*{\makecell[c]{Patch-wise \\ Representation \\ Discrepancy}} 
 & STAD \cite{STAD} & CVPR 2020 & 0.877 & 0.939 \\
 & PaDiM \cite{PaDiM} & ICPR 2020 & 0.955 & 0.975 \\
 \cline{2-5}
 & DFR \cite{DFR} & Neurocomputing 2021 & / & 0.950\\
 & FCDD \cite{FCDD} & ICLR 2021 & / & 0.920\\
  & MKD \cite{MKD} & CVPR 2021 & 0.877 & 0.907\\
 & Hou \emph{et al.} \cite{DivideAssemble} & ICCV 2021 & 0.895 & /\\
 & Metaformer \cite{Metaformer} & ICCV 2021 & 0.958 & /\\
 & DRAEM \cite{DRAEM} & ICCV 2021 & 0.980 & 0.973\\
 \cline{2-5}
  & RDAD \cite{RDAD} & CVPR 2022 & 0.985 & 0.978 \\
  & SSPCAB \cite{SSPCAB} & CVPR 2022 & 0.989 & 0.972 \\
   & DSR \cite{DSR} & ECCV 2022 & 0.982 & / \\
   & NSA \cite{NSA} & ECCV 2022 & 0.972 & 0.963 \\
   & UniAD \cite{UniAD} & NIPS 2022 & 0.966 & 0.966 \\ 
   & UTRAD \cite{UTRAD} & Neural Networks 2022 & 0.960 & 0.967\\
\hline
\hline
  \multirow{6}*{\makecell[c]{Patch-to-patch \\ Feature \\ Distance}} & PatchSVDD \cite{PatchSVDD} & ACCV 2020 & 0.921 & 0.957\\
  & DifferNet \cite{DifferNet} & WACV 2020 & 0.949 & /\\
  \cline{2-5}
 & CFLOW \cite{CFLOW} & WACV 2022 & 0.983 & \textcolor{red}{0.986}\\
 & CS-FLOW \cite{CS-FLOW} & WACV 2022 & 0.987 & /\\
 & Tsai \emph{et al.} \cite{CFLOW} & WACV 2022 & 0.981 & 0.981\\
 & PatchCore \cite{PatchCore} & CVPR 2022 & \textcolor{blue}{0.991} & 0.980 \\
\hline
\hline
\multirow{3}*{Others} & CutPaste \cite{CutPaste} & CVPR 2021 & 0.952 & 0.960 \\
  & Wang \emph{et al.} \cite{GlancingPatch} & CVPR 2021 & / & 0.91\\
 & SPD \cite{SPD} & ECCV 2022 & 0.946 & 0.946 \\
 \hline
 \hline
 \rowcolor{tableColor} Patch-wise\&Intra\&Inter & FOD (Ours) & - & \textcolor{red}{0.992} & \textcolor{blue}{0.983} \\
\hline
\end{tabular}}
\end{table}

\begin{table}
\caption{Detailed image-level AUROCs on the MVTecAD dataset.}
\label{tab:detailed_MVTecAD}
\resizebox{1.0\linewidth}{!}{
\begin{tabular}{c||cccccc}
\hline
\multirow{3}*{Category} & \multicolumn{6}{c}{Image-level Anomaly Detection} \\
\cline{2-7}
 & \makecell[c]{DRAEM \\ \cite{DRAEM}} & \makecell[c]{PatchSVDD \\ \cite{PatchSVDD}} & \makecell[c]{MKD \\ \cite{MKD}} & \makecell[c]{PatchCore \\ \cite{PatchCore}} & \makecell[c]{CFLOW \\ \cite{CFLOW}} & \makecell[c]{FOD \\ (Ours)}\\
\hline
\hline
 Carpet & 0.978 & 0.963 & \textbf{1.000} & \textbf{1.000} & 0.987 & \textbf{1.000}\\
  Grid  & \textbf{1.000} & 0.892 & 0.975 & 0.992 & 0.996 & \textbf{1.000}\\
  Leather & \textbf{1.000} & 0.953 & 0.956 & \textbf{1.000} & \textbf{1.000} & \textbf{1.000} \\
  Tile & 0.998 & 0.969 & 0.999 & \textbf{1.000} & 0.999 & \textbf{1.000}\\
  Wood  & \textbf{0.991} & 0.989 & 0.989 & 0.985 & \textbf{0.991} & \textbf{0.991}\\
\hline
  Bottle  & 0.993 & 0.976 & 0.989 & \textbf{1.000} & \textbf{1.000} & \textbf{1.000}\\
 Cable  & 0.929 & 0.899 & 0.972 & 0.992 & 0.976 & \textbf{0.995} \\
 Capsule & 0.984 & 0.763 & 0.979 & 0.984 & 0.977 & \textbf{1.000} \\
 Hazelnut & \textbf{1.000} & 0.912 & 0.997 & \textbf{1.000} & \textbf{1.000} & \textbf{1.000} \\
 Metal nut & 0.989 & 0.941 & 0.972 & \textbf{1.000} & 0.993 & \textbf{1.000} \\
 Pill & 0.981 & 0.791 & 0.971 & 0.954 & 0.968 & \textbf{0.984} \\
 Screw & 0.939 & 0.825 & 0.870 & 0.953 & 0.919 & \textbf{0.967} \\
 Toothbrush & \textbf{1.000} & 0.992 & 0.886 & 0.906 & 0.997 & 0.944 \\
 Transistor & 0.914 & 0.874 & 0.956 & 0.995 & 0.952 & \textbf{1.000} \\
 Zipper & \textbf{1.000} & 0.982 & 0.981 & 0.989 & 0.985 & 0.997 \\
\hline
\hline
 \rowcolor{tableColor} \textbf{Mean} & 0.980 & 0.915 & 0.966 & 0.983 & 0.983 & \textbf{0.992} \\
\hline
\end{tabular}}
\end{table}

\subsection{Main Results}
\label{sec:exp_anomaly_detection}
\textbf{SOTA Methods.} We extensively compare our method with those published SOTA methods in the past three years. The comparison results on MVTecAD are shown in Table \ref{tab:MVTecAD}. Then, we select five reproducible methods to report the detailed results on MVTecAD, these methods are representative and SOTA AD methods in the mainstream categories of image anomaly detection as we discuss in Related Work, including: DRAEM \cite{DRAEM}, PatchSVDD\cite{PatchSVDD}, PatchCore \cite{PatchCore}, MKD\cite{MKD}, CFLOW\cite{CFLOW}. For a fair comparison, we reproduce all these methods with the same backbone as in our model. Thus, despite using the unmodified code from the official repositories, we are not able to exactly reproduce the original results, but our numbers are very close. The detailed image-level AUROC results are shown in Table \ref{tab:detailed_MVTecAD}, and detailed pixel-level AUROC results are in the App. Table \ref{tab:sup_MVTecAD}. Most of the methods in Table \ref{tab:MVTecAD} don't report results on the BTAD and MVTec3D-RGB datasets, so we reproduce the five representative methods on the two datasets for comparison. The image-level and pixel-level AUROC results on BTAD and MVTec3D-RGB are shown in Table \ref{tab:BTAD_MVTec3D}. Additional detailed results are in the App. Table \ref{tab:sup_BTAD}, \ref{tab:sup_MVTecAD-3D}.

 \textbf{Anomaly Detection.} On MVTecAD, we set the SOTA performance on the mean detection AUROC, which is slightly higher than the best competitor, PatchCore \cite{PatchCore}. Note that the results in Table \ref{tab:detailed_MVTecAD} show that our method can achieve much better results than PatchCore when using the same backbone. What's more, in addition to the pill, screw and toothbrush classes, our method achieves more than 99\% AUROC in all other classes, while other methods only achieve more than 99\% AUROC in most nine classes. This shows that our method is more stable and effective in real-world applications. In the classes (\emph{e.g.} metal nut, screw, and transistor) with more global and logical anomalies, our method can achieve significantly better results than those methods depending on patch-wise discrepancy (\emph{e.g.} DRAEM \cite{DRAEM} and MKD \cite{MKD}), and can also achieve better results compared with other methods. This verifies that the three discrepancies are complementary factors and our model can simultaneously spot these discrepancies to recognize harder global and logical anomalies. On BTAD, our FOD can achieve 96.0\% mean detection AUROC, which can outperform the best competitor CFLOW \cite{CFLOW} by a margin of 1.2\%. On MVTec3D-RGB, we can outperform all previous SOTA methods by a margin of 3.3\%. Note that this dataset is much more challenging than the MVTecAD dataset when comparing the best results (99.2\% for MVTecAD vs. 88.4\% AUROC for MVTec3D-RGB). This demonstrates the robustness of our method. 

\begin{table}[t]
    \caption{Comparison of our method with the SOTA methods for image-level anomaly detection and pixel-level anomaly localization performance on the BTAD and MVTec3D-RGB datasets.}
\label{tab:BTAD_MVTec3D}
\resizebox{\linewidth}{!} {
\begin{tabular}{c||c|c|c|c|c||c}
\hline
\textbf{Method} & \makecell[c]{DRAEM \\ \cite{DRAEM}} & \makecell[c]{PatchSVDD \\ \cite{PatchSVDD}} & \makecell[c]{MKD \\ \cite{MKD}} & \makecell[c]{PatchCore \\ \cite{PatchCore}} & \makecell[c]{CFLOW \\ \cite{CFLOW}} & \makecell[c]{FOD \\ (Ours)} \\
\hline
\multicolumn{7}{c}{BTAD Dataset} \\
\hline
  \rowcolor{tableColor} \textbf{Image-level AUROC} &  0.922 & 0.924 & 0.935 & 0.934 & 0.948 & \textbf{0.960}\\
  \rowcolor{tableColor} \textbf{Pixel-level AUROC} &  0.942 & 0.964 & 0.965 & 0.976 & \textbf{0.978} & 0.975\\
\hline
\multicolumn{7}{c}{MVTec3D-RGB Dataset} \\
\hline
\rowcolor{tableColor} \textbf{Image-level AUROC} & 0.757 & 0.743 & 0.688 & 0.839 & 0.851 & \textbf{0.884}\\
  \rowcolor{tableColor} \textbf{Pixel-level AUROC} &  0.974 & 0.852 & 0.970 & \textbf{0.977} & 0.974 & 0.976\\
\hline
\end{tabular}}
\end{table}

\textbf{Anomaly Localization.} Our method can achieve comparable results with the best competitors on all three datasets. Our method is slightly lower than CFLOW \cite{CFLOW} and PatchCore \cite{PatchCore} on the three datasets, but we observe that our model that considers the intra- and inter-correlations outperforms the vanilla reconstruction AD models (\emph{e.g.} DRAEM \cite{DRAEM} and MKD \cite{MKD}) on all three datasets, which verifies the effectiveness of correlation modeling. Compared with patch-wise reconstruction methods, our method is more effective and robust to detect hard global and logical anomalies (see Figure \ref{fig:vis_anomalies}). 
%We select a subset containing these anomalies from MVTecAD and conduct experiments. The quantitative results are shown in the Table \ref{tab:MVTecAD}, and a qualitative comparison is shown in Figure \ref{fig:vis_anomalies}.  

\textbf{Qualitative Results.} Samples in Figure \ref{fig:motivation} show qualitative impressions of accurate anomaly localization from our method. It can be found that our approach can achieve the best anomaly score maps by combining the three discrepancies. Additional visualizations are in the App. Figure \ref{fig:sup_qualitative_results}.

\subsection{Ablation Study and Model Analysis}
\label{sec:ablation}
\textbf{Ablation Study.} To explain how our model works effectively, we further investigate the effect of the three key designs in our model: recognition views, entropy constraint and reference features. The quantitative results are shown in Table \ref{tab:ablation}, more results can be found in App. Table \ref{tab:sup_ablation}. The entropy constraint is quite effective and necessary in the intra-correlation branch. Specifically, it brings a remarkable 7.7\% averaged absolute AUROC promotion, which demonstrates that the entropy constraint strategy is really conducive to increase the distinguishability between abnormal and normal. Only utilizing pure reconstruction criterion or pure KL divergence can't get the most superior detection results, the combined criterion can outperform each single criterion consistently by a margin of 1.0\% and 9.8\%. Thus the reconstruction errors and the intra- and inter-correlations can collaborate to improve detection performance. For the external reference features, we compare the simple mean features with more elaborate coreset features \cite{PatchCore} (see App. Table \ref{tab:sup_reference_features} for more comparison methods). The results show that the simple mean features are more effective than the coreset features (0.923 vs. 0.836), which means that mean features are effective enough to represent all possible normal patterns and preserve the positional information. Finally, our proposed FOD surpasses the pure reconstruction Transformer by 4.0\% absolute improvement. These verify that our proposed explicit correlation learning approach is effective. 

\begin{table}[t]
    \caption{Ablation results in recognition views, entropy constraint, external reference features and anomaly scoring. \emph{Patch-wise}, \emph{Intra} and \emph{Inter} mean patch-wise discrepancy, intra- and inter-correlation, respectively. \emph{w/o} and \emph{w/} mean without and with entropy constraint. \emph{Mean} and \emph{Coreset} refer to mean and coreset features \cite{PatchCore} as the external reference features (see more details in App. \ref{sec:sup_reference_features}). \emph{Rec}, \emph{Div} and \emph{Rec\&Div} mean the pure reconstruction criterion, pure KL divergence (E.q.\ref{eq:KL_Div}) and the combined criterion (E.q.\ref{eq:combined}).}
\label{tab:ablation}
\resizebox{\linewidth}{!} {
\begin{tabular}{cccc||cccc}
\hline
\makecell[c]{Recognition \\ Views} & \makecell[c]{Entropy \\ Constraint} & \makecell[c]{Reference \\ Features} & \makecell[c]{Anomaly \\ Scoring} & MVTecAD & BTAD & MVTec3D-RGB & Avg\\
\hline
Patch-wise & / & / & Rec & 0.972 & 0.954 & 0.790 & 0.905\\
\cline{2-8}
\multirow{3}*{Intra} & w/o & / & Div & 0.700 & 0.811 & 0.708 & 0.740\\
 & w/ & / & Div & 0.911 & 0.822 & 0.717 & 0.817\\
 & w/ & / & Rec\&Div & 0.974 & 0.952 & 0.818 & 0.915\\
\cline{2-8}
\multirow{2}*{Inter} & w/ & Mean & Rec\&Div & 0.980 & 0.958 & 0.832 & 0.923\\
&  w/ & Coreset & Rec\&Div & 0.925 & 0.884 & 0.700 & 0.836\\
\cline{2-8}
Intra+Inter & w/ & Mean & Div & 0.896 & 0.922 & 0.814 & 0.877\\
\rowcolor{tableColor} \makecell[c]{\textbf{Patch-wise+Intra} \\ \textbf{+Inter (Ours)}} & w/ & Mean & Rec\&Div & \textbf{0.992} & \textbf{0.960} & \textbf{0.884} & \textbf{0.945}\\
\hline
\end{tabular}}
\end{table}

\begin{figure}
    \centering
    \includegraphics[width=1.0\linewidth]{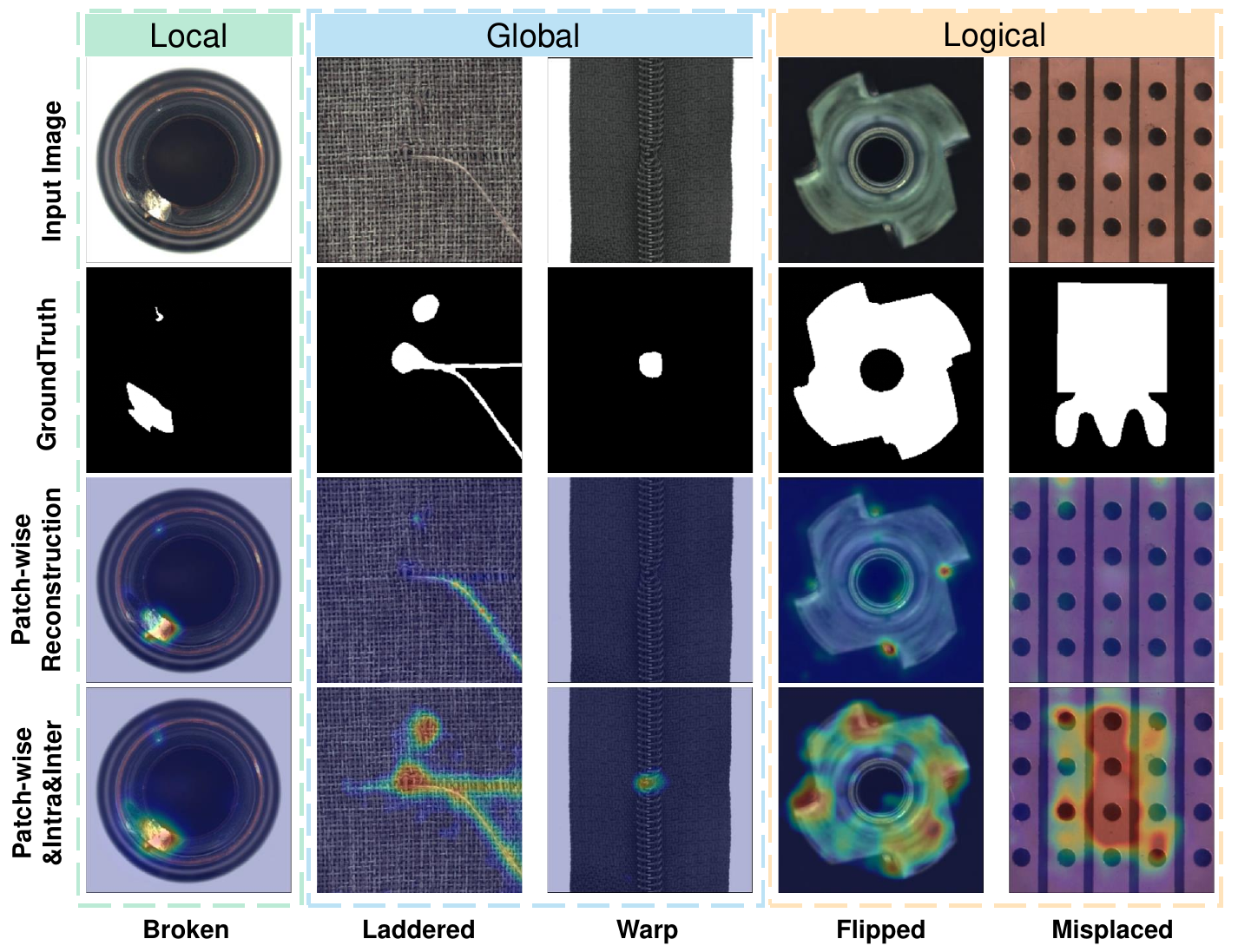}
    \caption{Visualization of different anomalies. The patch-wise reconstruction is only suitable for simple local anomalies, while for hard global and logical anomalies, it cannot detect or only detect partial anomalies. In contrast, our FOD can detect these anomalies more effectively.}
    \label{fig:vis_anomalies}
\end{figure}

\textbf{Effect for Different Anomalies.} To illustrate the effectiveness of our model intuitively, we explore the anomaly localization quality under different types of anomalies (see Figure \ref{fig:vis_anomalies}). \emph{E.g.}, simple local anomalies (Column 1) are obviously different from normal visuals; hard global anomalies (Columns 2,3) have less obvious visual appearances, which usually need to be effectively compared with normal regions to detect; logical anomalies (Columns 4,5) may be locally normal and can be detected correctly through the overall semantic understanding. We can find that our FOD is more distinguishable in general. For simple local anomalies, patch-wise reconstruction methods can also achieve good results. While for hard global anomalies, these methods cannot detect or only detect partial anomalies. However, for logical anomalies, these methods cannot detect them at all. In contrast, our method is conducive to detect diverse anomalies because of its complementary recognition views. This verifies that our method can make more precise detection and reduce the false-negative rate compared with the pure patch-wise reconstruction methods.

\begin{figure}
    \centering
    \includegraphics[width=1.0\linewidth]{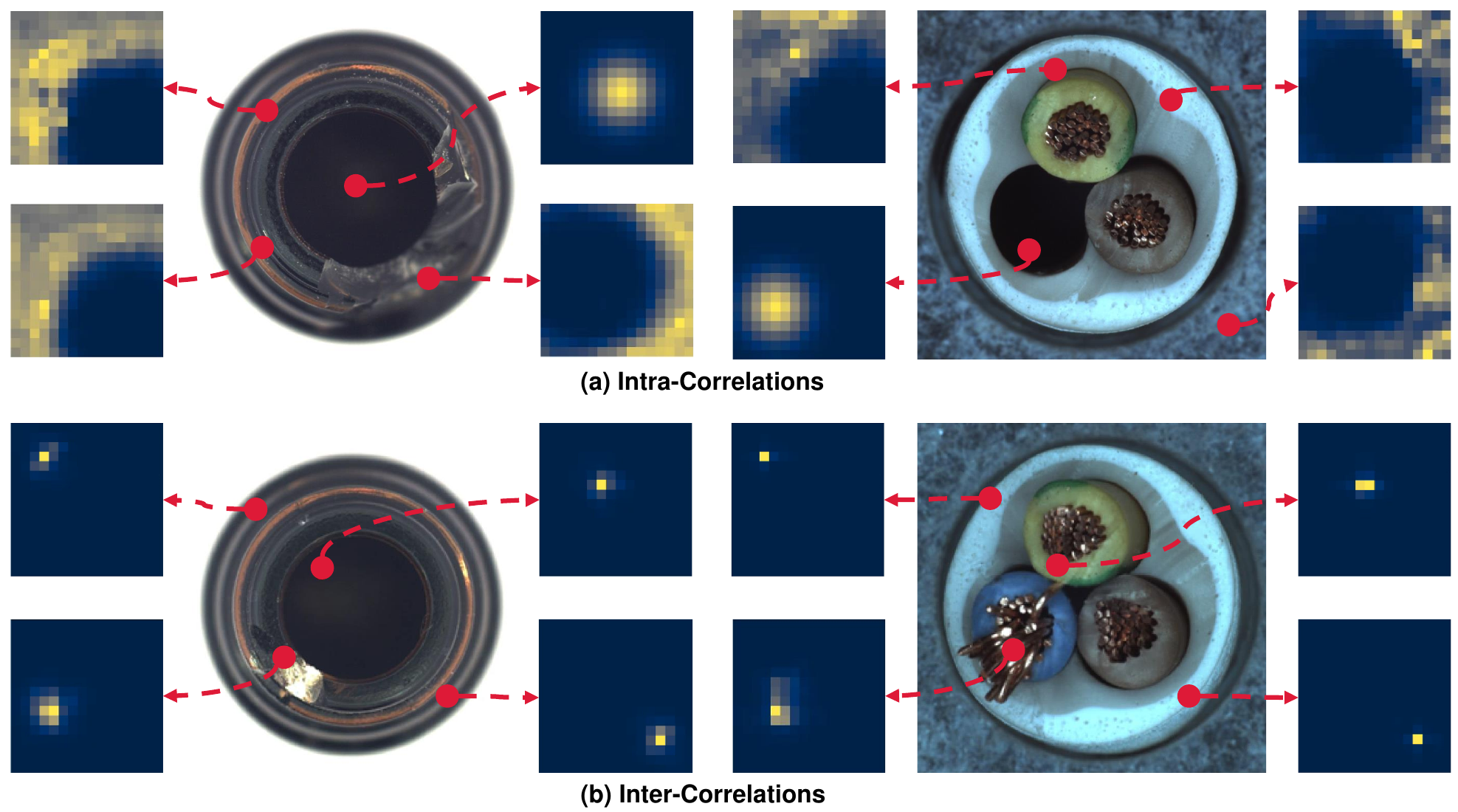}
    \caption{Visualization of intra- and inter-correlations. Each \textcolor{red}{red point} in the figure can represent an image patch, we plot its discrete correlation distribution with all the other image patches.}
    \label{fig:vis_correlations}
\end{figure}

\textbf{Correlations Visualization.} To explain what our model has learned intuitively, we visualize some learned correlations in Figure \ref{fig:vis_correlations} with the MVTecAD dataset. For the intra-correlation branch, the normal patches learn to build strong correlations with most patches in the whole image, while the correlation distributions of anomalies usually concentrate in the adjacent image patches. It can be found that the normal correlations are much spread and the abnormal correlations are more concentrated in the adjacent regions. For the inter-correlation branch, most normal correlations only concentrate in one point and the abnormal correlations are more spread, which exactly means that each normal patch can build strong correlations with one special normal pattern and abnormal patches are harder to establish correlations with normal patterns.

%------------------------------------------------------------------------
\section{Conclusion}
\label{sec:conclusion}
Humans recognize anomalies through two aspects: patch-wise representation discrepancies and weak patch-to-patch correlations. In this paper, we propose a novel AD framework: FOcus-the-Discrepancy, to simultaneously spot the patch-wise, intra- and inter-discrepancies. The patch-wise discrepancies and intra- and inter-correlations are complementary factors, and can be combined to detect more complex and diverse anomalies. The major characteristic of our method is that we renovate the self-attention maps in transformers to I2Correlation to explicitly establish intra- and inter-image correlations for AD modeling. The combination of explicit correlation learning and transformer architecture can match the core idea of anomaly detection quite well. 

\section*{Acknowledgements}
This work was supported in part by the National Natural Science Fund of China (61971281), the Shanghai Municipal Science and Technology Major Project (2021SHZDZX0102), and the Science and Technology Commission of Shanghai Municipality (22DZ2229005).

{\small
\bibliographystyle{ieee_fullname}
\bibliography{egbib}
}

\clearpage
\section*{Appendix}
\appendix
\label{sec:appendix}

\section{Implementation Details}
\subsection{Two-Phase Optimization Strategy}
\label{sec:sup_opt}

Note that directly optimizing the $\mathcal{L}_g$ (E.q.\ref{eq:loss_global}) will not make normal patches build strong correlations with most patches in the whole image. Instead, as both $\mathcal{T}^g$ and $\mathcal{S}^g$ have learnable parameters, it's easier to generate trivial solutions \cite{PRML}, where different patches have not learned position-adaptive $\mathcal{T}^g$, and $\mathcal{S}^g$ of normal and abnormal patches also collapse to a similar discrete distribution. Similarly, directly optimizing the $\mathcal{L}_e$ (E.q.\ref{eq:loss_external}) also can not make each normal patch establish a stronger correlation with a specific external normal pattern, the collapse issue may also exist. To better optimize the intra- and inter-correlation branches, we follow the minimax strategy in \cite{Anomaly-Transformer} to propose a two-phase optimization strategy. Specifically, in the first phase, we minimize the ${\rm Div}(\mathcal{T}^g, {\rm SG}[\mathcal{S}^g])$ item to make the target correlations adapt to various image patterns of different patches. In the second phase, we maximize the ${\rm Div}({\rm SG}[\mathcal{T}^g], \mathcal{S}^g)$ item to force the intra-correlations to pay more attention to the non-adjacent patches with the maximum entropy constraint. Through the two-phase optimization strategy, we can gradually distribute the weights in the intra-correlations to the non-adjacent patches, instead of all image patches forming similar intra-image patch-to-patch correlations, which can effectively reduce the risk of over-fitting. For the inter-correlation branch, the optimization goal is to make each normal patch establish a stronger correlation with a specific external normal pattern, so we first maximize the ${\rm Div}(\mathcal{T}^e, {\rm SG}[\mathcal{S}^e])$ item and then minimize the ${\rm Div}({\rm SG}[\mathcal{T}^e], \mathcal{S}^e)$, which is opposite to the optimization process in the intra-correlation branch. When optimizing intra- and inter-correlations, we also need to add the entropy constraint items:  maximizing ${\rm Ent}(\mathcal{S}^g)$ for the intra-correlation branch and minimizing ${\rm Ent}(\mathcal{S}^e)$ for the inter-correlation branch. Thus, combining the reconstruction loss $\mathcal{L}_{l}$, the loss functions of two phases are:

 \begin{align}
 \label{eg:sup_loss}
     \mathcal{L}_{1} = &\mathcal{L}_{l} + \lambda_1{\rm Div}(\mathcal{T}^g, {\rm SG}[\mathcal{S}^g]) -\lambda_1{\rm Div}(\mathcal{T}^e, {\rm SG}[\mathcal{S}^e]) \\ \nonumber
     \mathcal{L}_{2} = &\mathcal{L}_{l} - \lambda_1{\rm Div}({\rm SG}[\mathcal{T}^g], \mathcal{S}^g) + \lambda_1{\rm Div}({\rm SG}[\mathcal{T}^e], \mathcal{S}^e) \\ \nonumber 
     &-\lambda_2{\rm Ent}(\mathcal{S}^g) + \lambda_2{\rm Ent}(\mathcal{S}^e)
 \end{align}
where $\mathcal{L}_{l}$ is the reconstruction loss defined in E.q.\ref{eq:local_loss}, $SG[\cdot]$ means to stop gradient backpropagation, $\lambda_1$ and $\lambda_2$ are used to trade off the loss items. With the two-phase optimization strategy, each normal patch can establish stronger correlations with most normal patches and a stronger correlation with a specific external normal pattern, this is much harder for anomalies to achieve these correlations, thereby beneficial to amplify the normal-abnormal distinguishability. We further conduct ablation experiments on the direct optimization of $\mathcal{L}_g$ and $\mathcal{L}_e$ and the two-phase optimization strategy, the results are shown in Table \ref{tab:sup_strategy} of App. \ref{sec:sup_ablation}.

When implementing the two-phase optimization strategy, we can first calculate the backpropagated gradients of $\mathcal{L}_1$ loss and retain the gradient graph, and then calculate the backpropagated gradients of  $\mathcal{L}_2$ loss. The backpropagated gradients calculated in the second phase will be accumulated to the gradients in the first phase, and then we can call the optimizer to update the model parameters.

\subsection{Computation Cost}
 We provide computation cost analysis of our model and other compared models. All the values are measured with one NVIDIA RTX 3090 GPU and AMD EPYC 7453 28-Core CPU on the MVTecAD dataset, the results are shown in Table \ref{tab:sup_computation_cost}. For all models, we input a $256\times256$ image to calculate the FLOPs and set batch size to 4 to estimate the training time. Compared with other models, our model has the same order of magnitude of learnable parameters (PatchCore \cite{PatchCore} has no learnable parameters) and fewer FLOPs, but our model can achieve better detection results.
 
 \begin{table}
\begin{center}
\caption{Computation Cost Analysis of our model and other compared models.}
\label{tab:sup_computation_cost}
\resizebox{\linewidth}{!}{
\begin{tabular}{c|c|c|c|c}
\hline
 Method & FLOPs & \makecell[c]{Learnable \\ Parameters}  & \makecell[c]{Training Time \\ (one epoch)} & \makecell[c]{Inference \\ Speed} \\
 \hline
 \hline
 DRAEM \cite{DRAEM} & 198.7G & 97.4M & 15s & 22fps \\
 PatchSVDD \cite{PatchSVDD} & 23.6G & 15.3M & 16s & 1.2fps\\
  MKD \cite{MKD} & 24.1G & 24.9M & 10s & 23fps \\
 PatchCore \cite{PatchCore} & 12.1G & / & 14s & 19fps\\
  CFLOW \cite{CFLOW} & 30.7G & 24.7M & 74s & 9.5fps\\
  \hline
  \rowcolor{tableColor} FOD (Ours) & 10.9G & 16.2M & 15s & 21.4fps\\
\hline
\end{tabular}}
\end{center}
\end{table}

\section{Additional Results}
\subsection{Detailed Results}
\label{sec:sup_results}
The detailed pixel-level AUROC results of each category on the MVTecAD dataset are shown in Table \ref{tab:sup_MVTecAD}. The detailed results of each category for anomaly detection and localization performance on the BTAD and MVTec3D-RGB datasets are shown in Table \ref{tab:sup_BTAD} and \ref{tab:sup_MVTecAD-3D}. 

Table \ref{tab:sup_BTAD} shows the AUROCs of our method and the SOTA methods for detecting anomalies on the three classes of BTAD. Our FOD can achieve 96.0\% mean detection AUROC, which can outperform the best competitor CFLOW \cite{CFLOW} by
a margin of 1.2\%. 

The results for individual classes of MVTec3D-RGB are given in Table \ref{tab:sup_MVTecAD-3D}. We are able to outperform all previous SOTA methods regarding the average of all classes by a
margin of 3.3\%. Note that this dataset is more challenging than the MVTecAD dataset when comparing the best results (99.2\% for MVTecAD vs. 88.4\% AUROC for MVTec3D-RGB). Nevertheless, we detect defects in 6 out of 10 categories at an AUROC
of more than 90\%, while other methods only achieve
moth than 90\% AUROC in most four categories. This
demonstrates the robustness of our method.

\begin{table}
\caption{Detailed pixel-level AUROCs on the MVTecAD dataset.}
\label{tab:sup_MVTecAD}
\resizebox{1.0\linewidth}{!}{
\begin{tabular}{c||cccccc}
\hline
\multirow{3}*{Category} & \multicolumn{6}{c}{Pixel-level Anomaly Localization} \\
\cline{2-7}
 & \makecell[c]{DRAEM \\ \cite{DRAEM}} & \makecell[c]{PatchSVDD \\ \cite{PatchSVDD}} & \makecell[c]{MKD \\ \cite{MKD}} & \makecell[c]{PatchCore \\ \cite{PatchCore}} & \makecell[c]{CFLOW \\ \cite{CFLOW}} & \makecell[c]{FOD \\ (Ours)}\\
\hline
\hline
 Carpet & 0.955  & 0.953 & 0.990 & 0.991 & 0.994 & 0.990\\
  Grid & 0.997 & 0.961 & 0.986 & 0.988 & 0.993 & 0.989\\
  Leather & 0.986  & 0.978 & 0.978 & 0.994 & 0.997 &  0.995\\
  Tile & 0.992 & 0.911 & 0.952 & 0.948 & 0.969 & 0.948\\
  Wood  & 0.964 & 0.916 & 0.953 & 0.954 & 0.969 & 0.954\\
\hline
  Bottle & 0.991 & 0.978 & 0.985 & 0.989 & 0.988 & 0.987\\
 Cable & 0.947  & 0.964 & 0.972 & 0.985 & 0.975 & 0.986\\
 Capsule & 0.943 & 0.958 & 0.979 & 0.992 & 0.989 & 0.990\\
 Hazelnut & 0.997 & 0.978 & 0.982 & 0.986 & 0.984 & 0.989\\
 Metal nut & 0.995 & 0.980 & 0.972 & 0.980 & 0.971 &  0.985\\
 Pill & 0.976 & 0.963 & 0.971 & 0.963 & 0.976 & 0.986\\
 Screw & 0.976 & 0.957 & 0.983 & 0.994 & 0.988 & 0.992\\
 Toothbrush & 0.981 & 0.983 & 0.986 & 0.988 & 0.983 & 0.987\\
 Transistor & 0.909 & 0.970 & 0.886 & 0.968 & 0.923 & 0.989\\
 Zipper & 0.988 & 0.961 & 0.981 & 0.981 & 0.986 & 0.977\\
\hline
\hline
 \rowcolor{tableColor} \textbf{Mean} & 0.973 & 0.961 & 0.970 & 0.980 & 0.979 & \textbf{0.983}\\
\hline
\end{tabular}}
\end{table}

\begin{table}[t]
    \caption{Detailed comparison of our method with the SOTA methods for the image-level anomaly detection and pixel-level anomaly localization performance on the BTAD dataset.}
\label{tab:sup_BTAD}
\resizebox{\linewidth}{!} {
\begin{tabular}{c||cccccc}
\hline
\multicolumn{7}{c}{Image-level Anomaly Detection} \\
\hline
Category & \makecell[c]{DRAEM \\ \cite{DRAEM}} & \makecell[c]{PatchSVDD \\ \cite{PatchSVDD}} & \makecell[c]{MKD \\ \cite{MKD}} & \makecell[c]{PatchCore \\ \cite{PatchCore}} & \makecell[c]{CFLOW \\ \cite{CFLOW}} & \makecell[c]{FOD \\ (ours)} \\
\hline
Product01 & 0.995 & 0.984 & 0.938 & 0.984 & 1.000 & 0.995 \\
Product02 & 0.774 & 0.836 & 0.882 & 0.818 & 0.857 & 0.864\\
Product03 & 0.998 & 0.951 & 0.985 & 1.000 & 0.987 & 1.000\\
\rowcolor{tableColor} \textbf{Mean} & 0.922 & 0.924 & 0.935 & 0.934 & 0.948 & \textbf{0.960}\\
\hline
\multicolumn{7}{c}{Pixel-level Anomaly Localization} \\
\hline
Product01 & 0.927 & 0.948 & 0.949 & 0.973 & 0.971 & 0.971 \\
Product02 & 0.936 & 0.954 & 0.963 & 0.961 & 0.967 & 0.957\\
Product03 & 0.964 & 0.990 & 0.983 & 0.993 & 0.996 & 0.996\\
\rowcolor{tableColor} \textbf{Mean} & 0.942 & 0.964 & 0.965 & 0.976 & \textbf{0.978} & 0.975\\
\hline
\end{tabular}}
\end{table}

\begin{table*}
\caption{Detailed comparison of our method with the SOTA methods for image-level anomaly detection and pixel-level anomaly localization performance on the MVTec3D-RGB dataset.}
\centering
\label{tab:sup_MVTecAD-3D}
\resizebox{0.9\linewidth}{!}{
\begin{tabular}{c||cccccccccc|c}
\hline
\multicolumn{12}{c}{Image-level Anomaly Detection} \\
\hline
 Method & Bagel & Cable & Carrot & Cookie & Dowel & Foam & Peach & Potato & Rope & Tire & \textbf{Mean} \\
\hline
 DRAEM \cite{DRAEM} & 0.988 & 0.445 & 0.819 & 0.635 & 0.759 & 0.862  & 0.849 & 0.506 & 0.986 & 0.724 & 0.757 \\
 PatchSVDD \cite{PatchSVDD} & 0.892 & 0.831 & 0.570 & 0.695 & 0.722 & 0.626 & 0.618 & 0.653 & 0.999 & 0.827 & 0.743 \\
 MKD \cite{MKD} & 0.940 & 0.616 & 0.782 & 0.275 & 0.656 & 0.736 & 0.684 & 0.703 & 0.910 & 0.575 & 0.688 \\
 PatchCore \cite{PatchCore} & 0.887 & 0.939 & 0.903 & 0.703 & 0.972 & 0.809 & 0.750 & 0.581 & 0.959 & 0.884 & 0.839 \\
 CFLOW \cite{CFLOW} & 0.973 & 0.887 & 0.871 & 0.789 & 0.989 & 0.735 & 0.810 & 0.692 & 0.983 & 0.786 & 0.851 \\
 \rowcolor{tableColor} FOD (Ours) & 0.940 & 0.952 & 0.911 & 0.844 & 0.987 & 0.844 & 0.843 & 0.662 & 0.992 & 0.864 & \textbf{0.884}\\
\hline
\multicolumn{12}{c}{Pixel-level Anomaly Localization} \\
\hline
 DRAEM \cite{DRAEM} & 0.977 & 0.972 & 0.985 & 0.930 & 0.982 & 0.959  & 0.981 & 0.984 & 0.984 & 0.983 & 0.976\\
 PatchSVDD \cite{PatchSVDD} & 0.953 & 0.923 & 0.817 & 0.857 & 0.870 & 0.897 & 0.907 & 0.792 & 0.709 & 0.790 & 0.852\\
 MKD \cite{MKD} & 0.991 & 0.974 & 0.989 & 0.957 & 0.977 & 0.896  & 0.975 & 0.977 & 0.986 & 0.977 & 0.970\\
 PatchCore \cite{PatchCore} & 0.959 & 0.979 & 0.982 & 0.967 & 0.968 & 0.988 & 0.977 & 0.979 & 0.987 & 0.985 & \textbf{0.977}\\
 CFLOW \cite{CFLOW} & 0.984 & 0.982 & 0.984  & 0.974 & 0.987 & 0.900 & 0.982 & 0.983 & 0.980 & 0.981 & 0.974\\
 \rowcolor{tableColor} FOD (Ours) & 0.988 & 0.992 & 0.992 & 0.979 & 0.995 & 0.862 & 0.989 & 0.987 & 0.992 & 0.982 & 0.976\\
\hline
\end{tabular}}
\end{table*}

\subsection{Ablation Results}
\label{sec:sup_ablation}
Ablation results in pixel-level AUROC are shown in Table \ref{tab:sup_ablation}. The pixel-level AUROC results demonstrate the same conclusion as in Table \ref{tab:ablation}: the three key designs in our model: recognition views, entropy constraint, and reference features are all effective. These results also verify that our proposed explicit correlation learning approach is effective and the intra- and inter-image correlations are complementary factors to the patch-wise representation discrepancies.

\begin{table}[t]
    \caption{Ablation results in recognition views, entropy constraint, external reference features and anomaly scoring. \emph{Patch-wise}, \emph{Intra} and \emph{Inter} mean patch-wise discrepancy, intra- and inter-correlation, respectively. \emph{w/o} and \emph{w/} mean without and with entropy constraint. \emph{Mean} and \emph{Coreset} refer to mean and coreset features \cite{PatchCore} as the external reference features. \emph{Rec}, \emph{Div} and \emph{Rec\&Div} mean the pure reconstruction criterion, pure KL divergence (E.q.\ref{eq:KL_Div}) and the combined criterion (E.q.\ref{eq:combined}).}
\label{tab:sup_ablation}
\resizebox{\linewidth}{!} {
\begin{tabular}{cccc||ccc}
\hline
\makecell[c]{Recognition \\ Views} & \makecell[c]{Entropy \\ Constraint} & \makecell[c]{Reference \\ Features} & \makecell[c]{Anomaly \\ Scoring} & MVTecAD & BTAD & MVTec3D-RGB\\
\hline
Patch-wise & / & / & Rec & 0.974 & 0.975 & 0.964 \\
\cline{2-7}
\multirow{3}*{Intra} & w/o & / & Div & 0.717 & 0.602 & 0.778\\
 & w/ & / & Div & 0.804 & 0.620 & 0.863\\
 & w/ & / & Rec\&Div & 0.972 & 0.970 & 0.961\\
\cline{2-7}
\multirow{2}*{Inter} & w/ & Mean & Rec\&Div & 0.978 & \textbf{0.976} & 0.964\\
&  w/ & Coreset & Rec\&Div & 0.948 & 0.897 & 0.928\\
\cline{2-7}
Intra+Inter & w/ & Mean & Div & 0.846 & 0.831 & 0.965\\
\rowcolor{tableColor} \makecell[c]{\textbf{Patch-wise+Intra} \\ \textbf{+Inter (Ours)}} & w/ & Mean & Rec\&Div & \textbf{0.983} & 0.975 & \textbf{0.976}\\
\hline
\end{tabular}}
\end{table}

\textbf{Optimization Strategy.} Ablation results in optimization strategy are shown in Table \ref{tab:sup_strategy}. Directly optimizing the $\mathcal{L}_g$ and $\mathcal{L}_e$ cannot make the
intra-correlations pay more attention to the non-adjacent areas and will force the inter-correlations to pay more attention to diverse normal patterns. Moreover, direct optimization will cause the optimization problem of RBF kernel \cite{PRML}, thus cannot strongly amplify the difference between normal and abnormal patches as expected. The two-phase optimization strategy first optimizes the target-correlations to provide better guidance to the intra- and inter-correlations. Thus, the two-phase optimization strategy obtains more distinguishable correlation distributions than direct optimization and thereby performs better.

\begin{table}[t]
    \caption{Ablation results in optimization strategy. \emph{Direct} and \emph{Two-phase} mean direct optimization of $\mathcal{L}_g$ and $\mathcal{L}_e$ and the two-phase optimization strategy, respectively.}
\label{tab:sup_strategy}
\resizebox{\linewidth}{!} {
\begin{tabular}{c||cccccc}
\hline
Dataset & \multicolumn{2}{c}{MVTecAD} & \multicolumn{2}{c}{BTAD} & \multicolumn{2}{c}{MVTec3D-RGB}\\
\cline{2-7}
Strategy & Direct & Two-phase & Direct & Two-phase & Direct & Two-phase\\
\hline
\rowcolor{tableColor} \textbf{Image-level AUROC} & 0.986 & 0.991 & 0.959 & 0.960 & 0.828 & 0.884\\
\rowcolor{tableColor} \textbf{Pixel-level AUROC} & 0.957 & 0.982 & 0.956 & 0.975 & 0.952 & 0.976\\
\hline
\end{tabular}}
\end{table}

\begin{figure}
    \centering
    \includegraphics[width=1.0\linewidth]{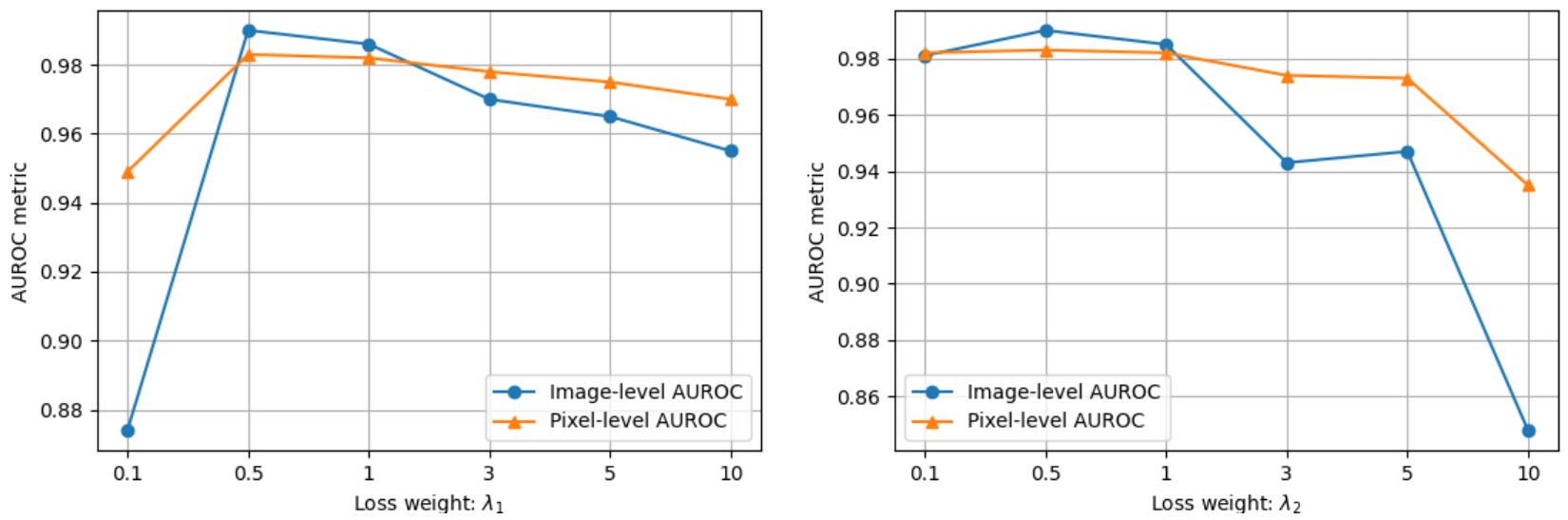}
    \caption{Hyper-parameter sensitivity for loss weights $\lambda_1$ and $\lambda_2$.}
    \label{fig:vis_hyperparameter}
\end{figure}

\textbf{Hyper-parameter Sensitivity.} We adopt the loss weights $\lambda_1$ and $\lambda_2$ to trade off the reconstruction loss, the correlation part and the entropy constraint part. The loss weight hyper-parameters $\lambda_1$ and $\lambda_2$ are set to $0.5$ and $0.5$ by default in the main text through comprehensive ablation experiments. To illustrate the sensitivity of our model, we further provide the model performance under different choices of the loss weights. Note that to avoid too many experiments, we only conduct experiments on the MVTecAD dataset, and fix $\lambda_2$ to $1$ to change $\lambda_1$ and then fix $\lambda_1$ to the best value to change $\lambda_2$. The ablation results are shown in Figure \ref{fig:vis_hyperparameter} and Table \ref{tab:sup_parameter_sensitivity}. It can be found that $\lambda_1$ and $\lambda_2$ are stable and easy to tune in the range of $0.5$ to $1$. The results verify that our model is not very sensitive to the loss weight hyper-parameters, which is essential for applications.

We also show hyper-parameter sensitivity for the number of heads and layers in Table \ref{tab:sup_heads} and \ref{tab:sup_layers}, respectively. It can be found that when setting the number of heads to $8$ and the number of layers to $3$ can achieve the best result. Thus, we use $8$ and $3$ as the default values in the main text.

\begin{table}[t]
    \caption{Hyper-parameter sensitivity for loss weights $\lambda_1$ and $\lambda_2$.}
\label{tab:sup_parameter_sensitivity}
\resizebox{\linewidth}{!} {
\begin{tabular}{c||cccccc}
\hline
 $\lambda_1$ & 0.1 & 0.5 & 1 & 3 & 5 & 10\\
\hline
\rowcolor{tableColor} \textbf{Image-level AUROC} & 0.874 & 0.990 & 0.986 & 0.970 & 0.965 & 0.955\\
\rowcolor{tableColor} \textbf{Pixel-level AUROC} & 0.949 & 0.983 & 0.982 & 0.978 & 0.975 & 0.970\\
\hline
\hline
$\lambda_2$ & 0.1 & 0.5 & 1 & 3 & 5 & 10\\
\hline
\rowcolor{tableColor} \textbf{Image-level AUROC} & 0.981 & 0.990 & 0.985 & 0.943 & 0.947 & 0.848\\
\rowcolor{tableColor} \textbf{Pixel-level AUROC} & 0.982 & 0.983 & 0.982 & 0.974 & 0.973 & 0.935\\
\hline
\end{tabular}}
\end{table}

\begin{table}[t]
    \caption{Hyper-parameter sensitivity for the number of heads.}
\label{tab:sup_heads}
\centering
\resizebox{0.65\linewidth}{!} {
 \begin{tabular}{c||ccc}
\hline
 Number of heads & 2 & 4 & 8 \\
\hline
\rowcolor{tableColor} \textbf{Image-level AUROC} & 0.987 & 0.990 & 0.992\\
\rowcolor{tableColor} \textbf{Pixel-level AUROC} & 0.982 & 0.983 & 0.983\\
\hline
\end{tabular}}
\end{table}

\begin{table}[t]
    \caption{Hyper-parameter sensitivity for the number of layers.}
\label{tab:sup_layers}
\centering
\resizebox{0.9\linewidth}{!} {
\begin{tabular}{c||ccccc}
\hline
 Number of layers & 2 & 3 & 4 & 5 & 6\\
\hline
\rowcolor{tableColor} \textbf{Image-level AUROC} & 0.986 & 0.990 & 0.983 & 0.979 & 0.957\\
\rowcolor{tableColor} \textbf{Pixel-level AUROC} & 0.982 & 0.983 & 0.978 & 0.978 & 0.969\\
\hline
\end{tabular}}
\end{table}

\textbf{Feature Levels.} Besides, we also explore the impact of different network layers on model performance and show the results in Table \ref{tab:sup_feature_levels}. For single-layer features, $8\times$ one-layer yields the best result as it trades off both semantic representation capability and fine-granularity of the features. Multi-scale feature fusion helps to improve the detection performance as it's conducive to cover more types and scales of anomalies. Note that using the $\{4\times, 8\times, 16\times\}$ three-layer features doesn't gain significant performance improvement compared with $\{8\times, 16\times\}$ two-layer features, but it instead increases the computational cost. Therefore, we use $\{8\times, 16\times\}$ two-layer features by default throughout the main text.

\begin{table}[t]
    \caption{Ablation results in feature levels. The experiments are conducted on the MVTecAD dataset. $4\times$, $8\times$, and $16\times$ mean feature maps with $\{4\times, 8\times, 16\times\}$ downsampling ratios, respectively.}
\label{tab:sup_feature_levels}
\resizebox{\linewidth}{!} {
\begin{tabular}{c||ccccc}
\hline
Feature Level & $4\times$ & $8\times$ & $16\times$ & $8\times$\&$16\times$ & $4\times$\&$8\times$\&$16\times$\\
\hline
\rowcolor{tableColor} \textbf{Image-level AUROC} & 0.885 & 0.981 & 0.975 & 0.990 & 0.988 \\
\rowcolor{tableColor} \textbf{Pixel-level AUROC} & 0.932 & 0.981 & 0.970 & 0.983 & 0.983 \\
\hline
\end{tabular}}
\end{table}

\subsection{External Reference Features}
\label{sec:sup_reference_features}
 External reference features are used for providing accumulated knowledge of normality for the inter-correlation learning branch. Thus, these features should represent all possible normal patterns of all normal samples. To this end, we can employ many methods to generate the external reference features, such as mean features, nearest features, sampling key features by coreset subsampling algorithm \cite{PatchCore}, generating prototype features by memory module \cite{MemoryAE}, and learning codebook features through vector quantization \cite{DSR} or sparse coding techniques \cite{Sparse}. However, because the RBF-kernel in $\mathcal{T}^e$ is position-sensitive, the reference features are better to preserve the positional information. In the following, we will introduce how to generate reference features in detail. 

\textbf{Mean Features.} Using patch-wise averaged features as the external reference features is really simple but effective. Formally, for position $(i, j)$, we first extract the set of patch features at $(i, j)$, $X_{ij} = \{x_{ij}^{k}\}, k \in [1, N]$ from the $N$ normal training images. Then, the reference features at position $(i, j)$ is computed as $x_{ij}^f = \frac{1}{N}\sum_{k=1}^{N}x_{ij}^k$. The final external reference features are composed of averaged features at all locations and then flattened into 1D: $X_f = {\rm Flatten}(\{x_{ij}^{f}\})$.

\textbf{Nearest Features.} To represent all possible normal patterns and also preserve the positional information, another simple way is to retain all normal features and then select the nearest features as the reference features. Specifically, we first extract the features of all images from the normal training set, which are denoted as $\mathcal{X} = \{X^k\}_{k=1}^N \in \mathbb{R}^{N\times d \times H \times W}$. Then, for each position $(i, j)$, we select its nearest normal feature in the $p \times p$ neighborhood as the reference feature $x_{ij}^f$. The $p \times p$ neighborhood is defined as follows:
\begin{align}
    \mathcal{N}_{(i,j)}^p = \{(i^\prime,j^\prime)|&i^\prime \in [i - \lfloor p/2 \rfloor,i + \lfloor p/2 \rfloor], \nonumber \\
    &j^\prime \in [j - \lfloor p/2 \rfloor,j + \lfloor p/2 \rfloor]\}
\end{align}
The reference feature $x_{ij}^f$ is calculated as follows:
\begin{equation}
    x_{ij}^f = \mathop{{\rm argmin}}\limits_{x \in \mathcal{X}_{(i.j)}^p}||x_{(i,j)} - x||_2
\end{equation}
where $\mathcal{X}_{(i.j)}^p = \{x_{(i^\prime,j^\prime)}^k|(i^\prime,j^\prime) \in \mathcal{N}_{(i,j)}^p, k = 1,2,\dots,N\}$ is the neighborhood features for position $(i, j)$, $x_{(i,j)}$ is the input feature at position $(i, j)$.

\textbf{Coreset Features.} Following \cite{PatchCore}, we can employ a coreset subsampling algorithm to sample key features as the reference features. The normal features $\mathcal{X} = \{X^k\}_{k=1}^N$ are also first extracted by a pre-trained network. Then, we can establish a coreset feature pool $\mathcal{X}^C$ by the coreset subsampling mechanism. Conceptually, coreset feature pool $\mathcal{X}^C$ aims to most closely and especially more quickly approximate the original features $\mathcal{X}$ in the feature space. Therefore, it can effectively preserve the key normal patterns in normal features. The \emph{minimax facility location coreset selection} algorithm is utilized, the procedure to generate $\mathcal{X}^C$ can be defined as follows:
\begin{equation}
    \mathcal{X}^{C*} = \mathop{{\rm argmin}}\limits_{\mathcal{X}^C \subset \mathcal{X}} \mathop{{\rm max}}\limits_{x_1 \in \mathcal{X}}
    \mathop{{\rm min}}\limits_{x_2 \in \mathcal{X}^C}||x_1 - x_2||_2
\end{equation}
The exact computation of $\mathcal{X}^{C*}$ is NP-Hard. We follow \cite{PatchCore} to use the iterative greedy approximation strategy to sample each coreset feature. The $i$th coreset feature $x^c_i$ in the coreset feature pool is sampled as follows:
\begin{equation}
\label{eq:subsampling}
    x^c_i \leftarrow \mathop{{\rm argmax}}\limits_{x \in \mathcal{X} - \mathcal{X}^C} \mathop{{\rm min}}\limits_{x^c \in \mathcal{X}^C}||x - x^c||_2 
\end{equation}
Then the coreset feature pool $\mathcal{X}^C$ is updated by $\mathcal{X}^C \leftarrow \mathcal{X}^C \cup \{x^c_i\}$. We can repeat the sampling process (E.q.\ref{eq:subsampling}) until the pre-defined coreset size.

\textbf{Prototype Features.} In MemoryAE \cite{MemoryAE}, the authors propose to use a memory module to generate prototype features of normal data for lessening the powerful reconstruction capability of CNNs to abnormal video frames. The memory module contains $P$ prototypes recording various prototypical patterns of normal data. However, the prototype features used in our method are slightly different, we need to learn $M$ prototype features at each location to preserve the position information. We denote prototype features at position $(i, j)$ by $\mathcal{P}_{(i, j)} = \{p_{(i,j )}^m\}_{m=1}^{M}$. We then perform the memory writing operation to update the prototype features.

To update each prototype feature $p_{(i, j)}^m$ at position $(i, j)$, we first need to select all input features declaring that the $p_{(i, j)}^m$ is the nearest one. Thus, we compute the cosine similarity between each input feature $x_{(i,j)}^k$ and all prototypes $\mathcal{P}_{(i, j)}$. The matching weights $w_{(i, j)}^{k,m}$ are as follows:
\begin{equation}
\label{eq:memory_matching}
    w_{(i, j)}^{k,m} = \frac{{\rm exp}(x_{(i,j)}^k(p_{(i, j)}^m)^T)}{\sum_{m^\prime=1}^{M}{\rm exp}(x_{(i,j)}^k(p_{(i, j)}^{m^\prime})^T)}
\end{equation}
 Note that multiple input features can be assigned to a single prototype in the memory. We denote by $U^m$ the set of indices for the corresponding input features for the $m$th item in the memory. We update the $m$th prototype using the input features indexed by the set $U^m$ as follows:
\begin{equation}
    p_{(i, j)}^m \leftarrow L_2(p_{(i, j)}^m + \sum\limits_{k \in U^m}\nu_{(i,j)}^{\prime,k,m}x_{(i, j)}^k)
\end{equation}
where $L_2$ means the $L2$ normalization. By using a weighted average of the input features, we can concentrate more on the input features close to the prototype. To this end, we can compute matching weights $\nu_{(i,j)}^{k,m}$ similar to E.q.\ref{eq:memory_matching}:
\begin{equation}
    \nu_{(i,j)}^{k,m} = \frac{{\rm exp}(x_{(i,j)}^k(p_{(i,j)}^m)^T)}{\sum_{k^\prime=1}^{K}{\rm exp}(x_{(i,j)}^{k^\prime}(p_{(i,j)}^{m})^T)}
\end{equation}
and renormalize it as follows:
\begin{equation}
    \nu_{(i,j)}^{\prime,k,m} = \frac{\nu_{(i,j)}^{k,m}}{{\rm max}_{k^\prime \in U^m}\nu_{(i,j)}^{k^\prime,m}}
\end{equation}

\textbf{Codebook Features.} Besides, we can also employ vector quantization (VQ) \cite{VQVAE} to learn codebook features as the reference features. Codebook features are highly semantic as VQ is based on quantizing the input features with features from a codebook $D \in \mathbb{R}^{N_e\times d}$ which has been trained for optimal decoding of spatial configurations of quantized features into high-fidelity images. For each input feature $x_{(i,j)}$ at position $(i, j)$, we can obtain a quantized feature representation $z_{(i,j)}$ by replacing the feature vector $x_{(i,j)}$ with its nearest neighbor $e_k$ in $D$:
\begin{equation}
\label{eq:quantization}
   z_{(i,j)} = e_k, \quad where \;\; k = \mathop{{\rm argmin}}\limits_j||x_{(i,j)} - e_j||_2
\end{equation}
After quantizing the input features to the codebook features, we feed the quantized features to a decoder. The decoder output feature $o_{(i,j)}$ at position $(i, j)$ aims at reconstructing the input feature $x_{(i,j)}$. During learning the codebook features, we maximize the cosine similarity between the decoder output $o_{(i,j)}$ and the input $x_{(i,j)}$. Note that the quantization process (E.q.\ref{eq:quantization}) is non-differentiable, but we could approximate the gradient similar to the straight-through estimator and directly copy gradients from decoder input $z_{(i,j)}$ to input feature $x_{(i,j)}$ \cite{VQVAE}. The learning objective is defined as:
\begin{align}
    {\rm max}\sum_{i=1}^{H}&\sum_{j=1}^{W}{\rm cos}(o_{(i,j)},x_{(i,j)}) \nonumber  \\
    &- ||sg[x_{(i,j)}]-e_k||_2 - ||x_{(i,j)}-sg[e_k]||_2
\end{align}
With the codebook features, we can use the quantized feature $z_{(i,j)}$ as the reference feature $x_{ij}^f$ as position $(i,j)$.

\begin{table}[t]
    \caption{Ablation results in external reference features. \emph{Mean}, \emph{Nearest}, \emph{Coreset}, \emph{Prototype}, and \emph{Codebook Features} refer to mean, nearest, coreset \cite{PatchCore}, prototype \cite{MemoryAE}, and codebook \cite{DSR} features as the external reference features, respectively. $\cdot$/$\cdot$ means image-level and pixel-level AUROCs, respectively.}
\label{tab:sup_reference_features}
\resizebox{\linewidth}{!} {
\begin{tabular}{c||ccc}
\hline
\multirow{2}*{\makecell[c]{Reference \\ Features}} & \multicolumn{3}{c}{Dataset}\\
\cline{2-4}
   & MVTecAD & BTAD & MVTec3D-RGB\\
\hline
\rowcolor{tableColor} Mean Features & 0.990/0.983 & 0.960/0.975 & 0.884/0.976\\
\rowcolor{tableColor} Nearest Features & 0.973/0.969 & 0.947/0.973 & 0.842/0.970\\
\rowcolor{tableColor} Coreset Features \cite{PatchCore} & 0.925/0.948 & 0.884/0.897 & 0.700/0.928\\
\rowcolor{tableColor} Prototype Features \cite{MemoryAE} & 0.987/0.978 & 0.958/0.975 & 0.898/0.982\\
\rowcolor{tableColor} Codebook Features \cite{DSR} & 0.970/0.970 & 0.955/0.970 & 0.797/0.956\\
\cline{2-4}
\hline
\end{tabular}}
\end{table}

\textbf{Results.} Ablation results in external reference features are shown in Table \ref{tab:sup_reference_features}. As we mentioned, the reference features are better to preserve the positional information, the results also show that the methods (\emph{e.g.}, Coreset Features and Codebook Features) that can't preserve the position information performs worse. Prototype features can achieve comparable performance with mean features, but it's more intricate to generate prototype features by memory module \cite{MemoryAE}. The ablation results demonstrate that although the mean features are simple, they are quite effective and can achieve better results than these more intricate reference feature generation methods. 

\section{Qualitative Results}
\begin{figure*}
    \centering
    \includegraphics[width=0.7\linewidth]{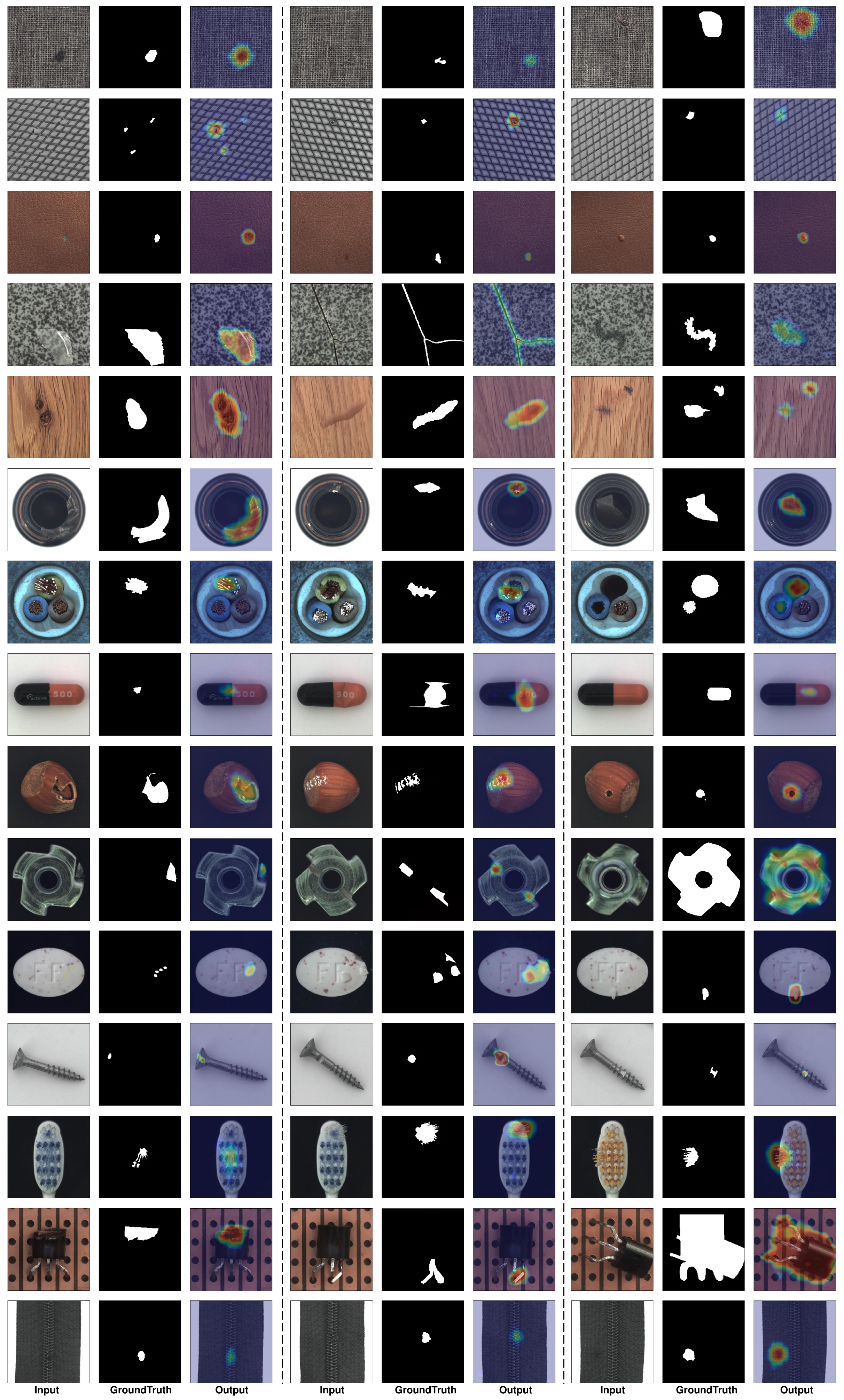}
    \caption{Visualization of anomaly localization maps generated by our method on industrial inspection data. All examples are from the MVTecAD dataset.}
    \label{fig:sup_qualitative_results}
\end{figure*}

We present in Figure \ref{fig:sup_qualitative_results} additional anomaly localization results of categories with different anomalies in the MVTecAD dataset.

\end{document}